\documentclass[accepted]{article}

\usepackage{microtype}
\usepackage{graphicx}
\usepackage{subfigure}
\usepackage{booktabs} 

\usepackage{hyperref}
\usepackage{xurl}
\usepackage{comment}

\usepackage{icml2023}

\usepackage{amsmath}
\usepackage{amssymb}
\usepackage{mathtools}
\usepackage{amsthm}

\usepackage[capitalize,noabbrev]{cleveref}

\theoremstyle{plain}

\theoremstyle{definition}

\theoremstyle{remark}

\usepackage[textsize=tiny]{todonotes}

\icmltitlerunning{Eight Things to Know about Large Language Models}

\begin{document}

\twocolumn[
\icmltitle{Eight Things to Know about Large Language Models}



\icmlsetsymbol{equal}{*}

\begin{icmlauthorlist}
\icmlauthor{Samuel R. Bowman}{nyu,ant}
\end{icmlauthorlist}

\icmlaffiliation{nyu}{New York University}
\icmlaffiliation{ant}{Anthropic, PBC}

\icmlcorrespondingauthor{Samuel R. Bowman}{bowman@nyu.edu}

\icmlkeywords{Machine Learning, ICML}

\vskip 0.3in
]



\printAffiliationsAndNotice{} 

\begin{abstract}
The widespread public deployment of large language models (LLMs) in recent months has prompted a wave of new attention and engagement from advocates, policymakers, and scholars from many fields. This attention is a timely response to the many urgent questions that this technology raises, but it can sometimes miss important considerations. This paper surveys the evidence for eight potentially surprising such points:
\begin{enumerate}
\item LLMs predictably get more capable with increasing investment, even without targeted innovation.
\item Many important LLM behaviors emerge unpredictably as a byproduct of increasing investment.
\item LLMs often appear to learn and use representations of the outside world.
\item There are no reliable techniques for steering the behavior of LLMs.
\item Experts are not yet able to interpret the inner workings of LLMs.
\item Human performance on a task isn’t an upper bound on LLM performance.
\item LLMs need not express the values of their creators nor the values encoded in web text.
\item Brief interactions with LLMs  are often misleading.
\end{enumerate}
\end{abstract}

\section*{Introduction}

Large language models \citep[LLMs, e.g. GPT-3, PALM, LLaMA, and GPT-4;][]{brown2020gpt3,chowdhery2022palm,touvron2023llama,openai2023gpt4} and products built on them, such as ChatGPT, have recently prompted an enormous amount of attention from journalists, \cite{klein2023this,billyperrigo,lastweektonight}, policymakers \cite{paul2023chatgpt,diane2023as,lieu2023Im}, and scholars from many fields \cite{chan2022gpt,lund2023chatting,choi2023chatgpt,biswas2023chatgpt}. This technology defies expectations in many ways, though, and it can be easy for brief discussions of it to leave out important points.

This paper presents eight potentially surprising claims that I expect will be salient in at least some of the conversations that are springing up around LLMs. They reflect, to the best of my understanding, views that are reasonably widely shared among the researchers---largely based in private labs---who have been developing these models. All the evidence I present here, as well as most of the arguments, are collected from prior work, and I encourage anyone who finds these claims useful to consult (and directly cite) the sources named here.

I do not mean for these claims to be normative in any significant way. Rather, this work is motivated by the recognition that deciding what we should do in light of this disruptive new technology is a question that is best led—in an informed way—by scholars, advocates, and lawmakers from outside the core technical R\&D community.

\section{LLMs predictably get more capable with increasing investment, even without targeted innovation}

\begin{figure*}
    \centering
\includegraphics[scale=0.29]{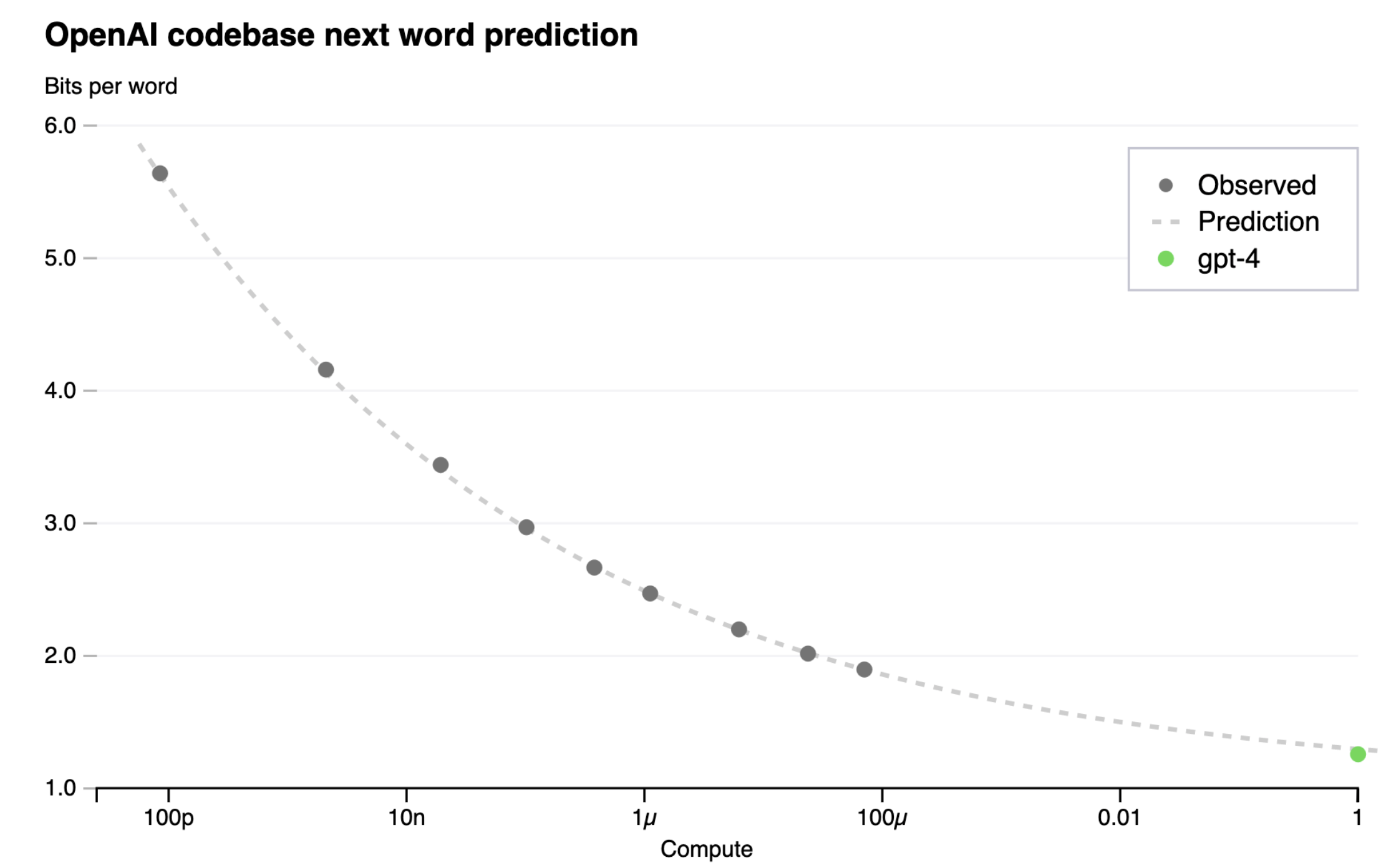}
    \caption{Excerpted from \citet{openai2023gpt4}: A scaling law result for one measure of language model performance, showing a consistent trend as the amount of computation used to train a model is scaled up 10,000,000,000$\times$
 times from a small prototype system to GPT-4.}
    \label{fig:OpenAI_codebase_next_word}
\end{figure*}

\textit{Scaling law} results \cite{kaplan2020scaling,brown2020gpt3,hoffmann2022chinchilla} have been a major driving factor in the recent surge of research and investment into LLMs \cite{ganguli2022predictability}. Scaling laws allow us to precisely predict some coarse-but-useful measures of how capable future models will be as we scale them up along three dimensions: the amount of data they are fed, their size (measured in parameters), and the amount of computation used to train them (measured in FLOPs). These results thereby allow us to make some key design decisions, such as the optimal size of a model given some fixed resource budget, without extremely expensive trial and error.

Our ability to make this kind of precise prediction is unusual in the history of software and unusual even in the history of modern AI research. It is also a powerful tool for driving investment since it allows R\&D teams to propose model-training projects costing many millions of dollars, with reasonable confidence that these projects will succeed at producing economically valuable systems.

Concretely, consider these three superficially very different systems: OpenAI’s original GPT can perform simple text-labeling tasks but cannot generally produce coherent text \cite{radford2018gpt1}. GPT-2 adds the ability to produce text of reasonably high quality, as well as a limited ability to follow simple instructions \cite{radford2019gpt2}. GPT-3 is the first modern general-purpose LLM, and is practically useful across a wide range of language tasks. The designs of these three models hardly differ at all. Instead, the qualitative differences between them stem from vast differences in scale: Training GPT-3 used roughly 20,000$\times$ more computation than training the original GPT \cite{sevilla2022epoch}, as well as significantly more data and parameters. There \textit{are} substantial innovations that distinguish these three models, but they are almost entirely restricted to infrastructural innovations in high-performance computing rather than model-design work that is specific to language technology. 

While the techniques used to train the newest LLMs are no longer generally disclosed, the most recent detailed reports suggest that there have been only slight deviations from this trend, and that designs of these systems are still largely unchanged \cite{chowdhery2022palm, hoffmann2022chinchilla,touvron2023llama}.

Continuing to scale these techniques up beyond GPT-3 has produced further economically valuable returns: The subsequent GPT-4 model outperforms qualified humans on many graduate and professional exams \cite{openai2023gpt4}, and its development helped prompt a multi-billion-dollar investment in the company that built it \cite{ashley2023microsoft}. Scaling laws allowed the creators of GPT-4 to cheaply and accurately predict a key overall measure of its performance: This forecast was made by fitting a statistical trend in the performance of small models, which collectively took about 0.1\% of the resources needed by the final model, and then extrapolating out that trend (see \cref{fig:OpenAI_codebase_next_word}).

\section{Specific important behaviors in LLM tend to emerge unpredictably as a byproduct of increasing investment\label{sec:unpredictability}}

\begin{figure*}
    \centering
    \includegraphics[scale=0.35]{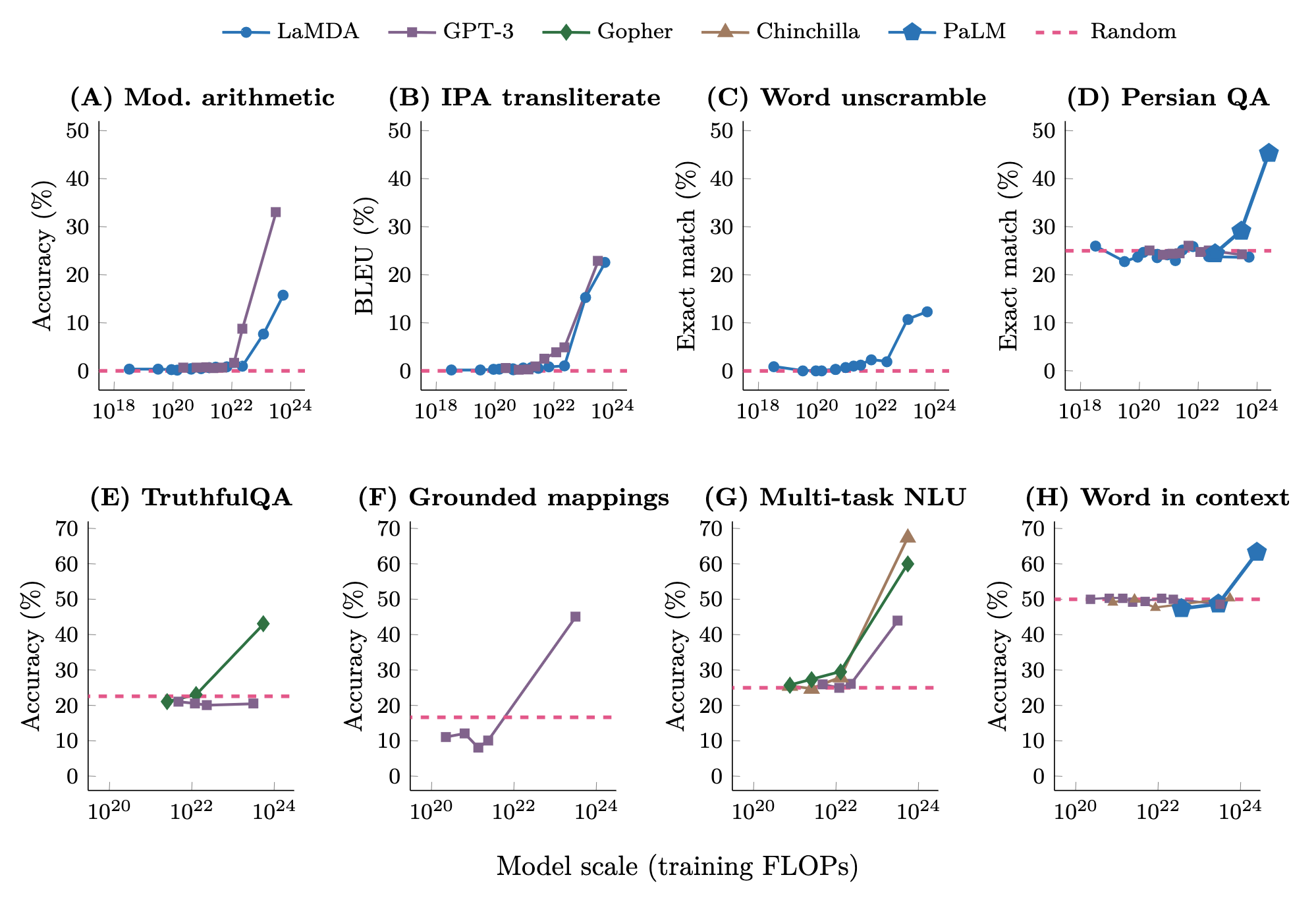}
    \caption{Excerpted from \citet{wei2022emergent}: Evaluations of performance on specific tasks or behaviors in LLMs do not generally show predictable trends, and it is common for new behaviors to emerge abruptly when transitioning from a less resource-intensive version of a model to a more resource-intensive one.}
    \label{fig:wei2022emergent}
\end{figure*}

Scaling laws generally only predict a model’s \textit{pretraining test loss}, which measures the model's ability to correctly predict how an incomplete piece of text will be continued.\footnote{Much of the data and computer time that goes into building a modern LLM is used in an expensive initial \textit{pretraining} process. Language-model pretraining intuitively resembles the autocomplete task: In it, an artificial neural network model takes in a text one word at a time, makes a probabilistic prediction about which word will come next, and has its behavior incrementally adjusted to make it assign a greater probability to the actual next word in similar contexts in the future. Pretraining test loss measures how effectively an LLM has learned to make these predictions.}
While this measure is correlated with how useful a model will be on average across many practical tasks \cite{radford2019gpt2}, it is largely \textit{not} possible to predict when models will start to show specific skills or  become capable of specific tasks \citep[see \cref{fig:wei2022emergent};][]{Steinhardt2021risks,ganguli2022predictability,wei2022emergent}. Often, a model can fail at some task consistently, but a new model trained in the same way at five or ten times the scale will do well at that task.

\begin{figure*}
    \centering
    \includegraphics[scale = 0.35]{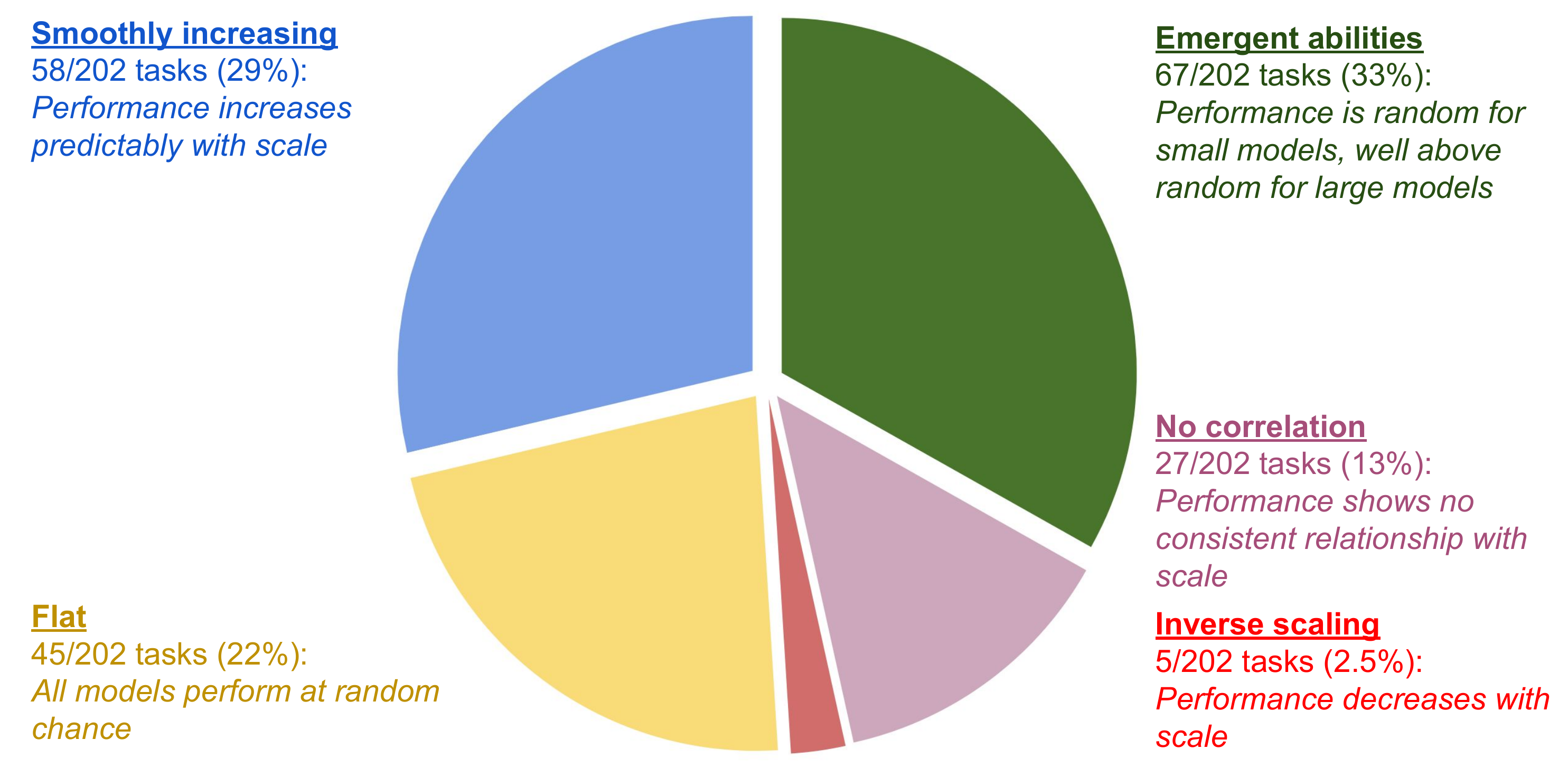}
    \caption{Adapted from a figure by Jason Wei based on data from \citet{wei2022emergent}: The 202 tasks evaluated in the language-technology benchmark BIG-Bench \cite{srivastava2022beyond} tend to show improved performance with scale overall, but individually they can improve gradually, improve abruptly, stay level, get worse, or vacillate, making it impossible to extrapolate the performance of some future system confidently.}
    \label{fig:big_bench}
\end{figure*}

\citeauthor{wei2022emergent} show that the tasks in BIG-Bench \cite{srivastava2022beyond}, a standard broad-coverage benchmark for LLM abilities, show a range of different kinds of trend that collectively make scaling-law style predictions unreliable (\cref{fig:big_bench}). This means that when a lab invests in training a new LLM that advances the scale frontier, they’re \textit{buying a mystery box}: They’re justifiably confident that they’ll get a variety of economically valuable new capabilities, but they can make few confident predictions about what those capabilities will be or what preparations they’ll need to make to be able to deploy them responsibly.

Concretely, two of the key behaviors in GPT-3 that set it apart as the first modern LLM are that it shows \textit{few-shot learning}, the ability to learn a new task from a handful of examples in a single interaction, and \textit{chain-of-thought reasoning}, the ability to write out its reasoning on hard tasks when requested, as a student might do on a math test, and to show better performance as a result. GPT-3’s capacity for few-shot learning on practical tasks appears to have been discovered only after it was trained, and its capacity for chain-of-thought reasoning was discovered only several months after it was broadly deployed to the public \cite{nye2021show,wei2022chain,kojima2022large,zhou2023least}.\footnote{See \citet{branwen} for a survey that includes additional unpublished reports of this behavior.} 
In addition, model abilities involving programming, arithmetic, defusing misconceptions, and answering exam questions in many domains show abrupt improvements as models are scaled up \cite{wei2022emergent,srivastava2022beyond}.

There are few widely agreed-upon limits to what capabilities could emerge in future LLMs. While there are some hard constraints on the behaviors of typical current LLMs---stemming from limits on the amount of text they can use as input at any one time, limits on their ability to interact with the world during training, or limits on the amount of computation they can perform for each word they generate---it is arguably plausible that these will be overcome with further research within the same technical paradigm. However, many experts disagree:  51\% of language-technology researchers surveyed in spring 2022 agreed that “expert-designed strong inductive biases (à la universal grammar, symbolic systems, or cognitively-inspired computational primitives) will be necessary to practically solve some important real-world problems or applications in [language technology]”, which if true would represent a limit to the LLM paradigm \cite{michael2022nlp}.

Expert forecasts, however, have often predicted that we would see less progress with LLMs than has actually occurred. While forecasts by technology researchers are often informal, and I am aware of no precise evaluation of their accuracy, we do have a crisp example of experienced professional forecasters making similar mistakes: \citet{Steinhardt2021one} presents results from a competition that was organized in summer 2021, which gave forecasters access to experts, extensive evidence, and a cash incentive, and asked them to predict what state-of-the-art performance with LLMs would be in each of the next four years on two specific tasks. The results from summer 2022, only one year into the competition, substantially exceeded what the consensus forecast said would be possible in \textit{2024}. Results with GPT-4 in early 2023 exceeded the consensus forecast for \textit{2025} on the one measure for which we have reported results \cite{openai2023gpt4}. This suggests that it is worth planning for the possibility that we continue to see fast technical progress. 

\section{LLMs often appear to learn and use representations of the outside world\label{sec:LLMS_world_models}}

There is increasingly substantial evidence that LLMs develop internal representations of the world to some extent, and that these representations allow them to reason at a level of abstraction that is not sensitive to the precise linguistic form of the text that they are reasoning about. Current LLMs seem to do this only weakly and sporadically, but the evidence for this phenomenon is clearest in the largest and most recent models, such that we should expect it to become more robust as systems are scaled up further.

\begin{figure}
    \centering
    \includegraphics[scale = 0.2]{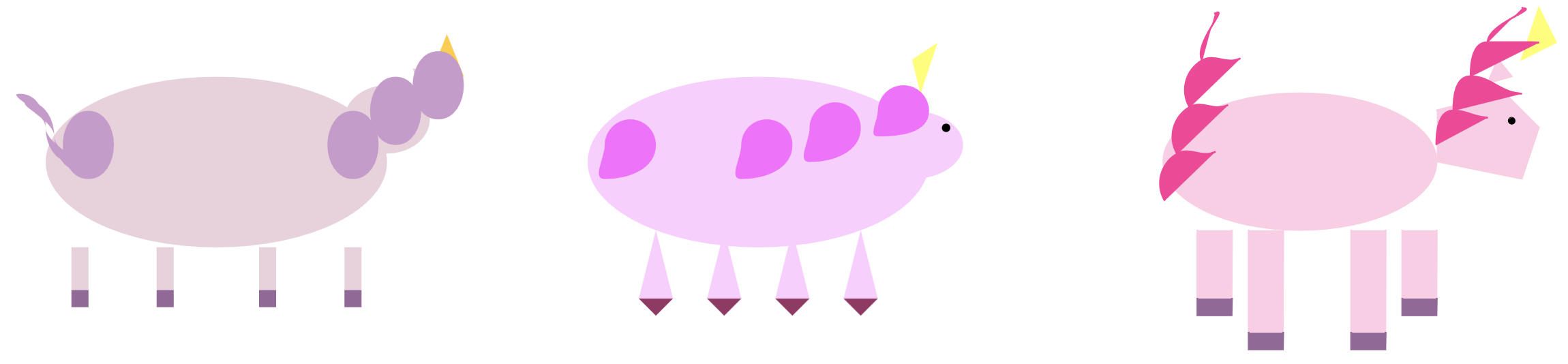}
    \caption{Excerpted from \citet{bubeck2023sparks}: An popular informal (and potentially cherry-picked) demonstration of LLMs’ ability to manipulate visual representations. Here, a private version of GPT-4, trained without any access to visual information, is asked to write instructions in a graphics programming language to draw a unicorn. During the model’s training (left to right), the resulting drawings appear to become more competent.}
    \label{fig:three_unicorns}
\end{figure}

Evidence for this claim includes the following results, spanning many established experimental methods and models:
\begin{itemize}
    \item Models’ internal representations of color words closely mirror objective facts about human color perception \cite{abdou2021can,patel2022mapping,sogaard2023grounding}.
    \item Models can make inferences about what the author of a document knows or believes and use these inferences to predict how the document will be continued \cite{andreas2022language}.
    \item Models use internal representations of the properties and locations of objects described in stories, which evolve as more information about these objects is revealed \cite{li2021implicit}. This can include the ability to internally represent the spatial layout of the setting of a story \cite{patel2022mapping,bubeck2023sparks}. Models also use similar representations for facts about real-world geography \cite{lietard2021language}. 
    \item Models can at least sometimes give instructions describing how to draw novel objects \citep[Figure \ref{fig:three_unicorns};][]{bubeck2023sparks}.
    \item Models that are trained to play board games from descriptions of individual game moves, without ever seeing a full depiction of the game board, learn internal representations of the state of the board at each turn \cite{li2023emergent}.
    \item Models can distinguish common misconceptions from true facts \cite{wei2022emergent}, and often show  well-calibrated internal representations for how likely a claim is to be true \cite{kadavath2022language,wei2022emergent,burns2022latent_knowledge}.
    \item Models pass many tests designed to measure commonsense reasoning, including some like the Winograd Schema Challenge that are explicitly designed to include no purely textual clues to the answer \cite{levesque2012winograd,he2021deberta,openai2023gpt4}.
\end{itemize}

These results are in tension, at least to some extent, with the common intuition that LLMs \textit{are nothing but statistical next-word predictors}, and therefore cannot learn or reason about anything but text. While the premise of this intuition is technically correct in some cases, it can paint a misleading picture of the often-rich representations of the world that LLMs develop as they are trained. In addition, LLMs are increasingly often augmented with other ways of learning about the world that make this claim literally false, such as through interactive training methods \cite{ziegler2019fine,stiennon2020rlhf,ouyang2022InstructGPT,bai2022training}, integration with image processing systems, \cite{alayrac2022flamingo,openai2023gpt4}, or integration with other software tools  \cite{nakano2021webgpt,menick2022gopher,collins2022structured,schick2023toolformer,openai2023ChatGPTplugins}.

\section{There are no reliable techniques for steering the behavior of LLMs\label{sec:control_and_robustness}}

Much of the expense of developing an LLM goes into language-model pretraining: The process of training a neural network to predict how random samples of human-written text will be continued. In most cases, though, the developers of such a system want to use it for tasks other than predicting continuations, which requires that it be adapted or guided in some way. Even building a general-purpose \textit{instruction-following} model, where one is not attempting to specialize on any particular task, requires this kind of adaptation: Otherwise, the model will attempt to \textit{continue} its instructions rather than following them \citep{ouyang2022InstructGPT}. 

This adaptation typically involves one or more of these three techniques:

\begin{enumerate}
    \item Plain language model prompting, where one prepares an incomplete text like ``\textit{The translation of ‘cat’ in French is`}'', such that a typical continuation of the text should represent a completion of the intended task \cite{radford2019gpt2,raffel2020T5}.\footnote{Prompting in the more general sense can describe the practice of writing instructions or requests for an LLM, where these instructions and requests need not have this continuation property. The base models produced by language-model pretraining do not support this kind of prompting.}
    \item Supervised fine-tuning, where one trains the model to match high-quality human demonstrations on the task \cite{radford2018gpt1,devlin2018bert,ouyang2022InstructGPT}.
    \item Reinforcement learning, where one incrementally weakens or strengthens certain model behaviors according to preference judgments from a human tester or user \cite{ziegler2019fine,stiennon2020rlhf,ouyang2022InstructGPT,bai2022training}.
\end{enumerate}

These techniques produce useful systems, but they are far from perfectly effective: They can't guarantee that an AI model will behave appropriately in every plausible situation it will face in deployment. Nor can they even make a model \textit{try} to behave appropriately to the extent possible given its skills and knowledge (to the extent that it can be said to have generalizable skills or knowledge). In particular, models can misinterpret ambiguous prompts or incentives in unreasonable ways, including in situations that appear unambiguous to humans, leading them to behave unexpectedly \cite{damour2020underspecification,kenton2021alignment}.

In one key way, this problem is getting easier to tackle: As LLMs become more capable of using human language and human concepts, they also become more capable of learning the generalizations we would like. Indeed, many control techniques work better with larger models, at least for simple tasks\cite{hendrycks2020pretrained,bai2022training,chung2022scaling,ganguli2023capacity}. In another important way, though, the problem is becoming more difficult: More capable models can better recognize the specific circumstances under which they are trained. Because of this, they are more likely to learn to act as expected in precisely those circumstances while  behaving competently but unexpectedly in others. This can surface in the form of problems that \citet{perez2022discovering} call \textit{sycophancy}, where a model answers subjective questions in a way that flatters their user’s stated beliefs, and \textit{sandbagging}, where models are more likely to endorse common misconceptions when their user appears to be less educated. It seems likely that issues like these played some role in the bizarre, manipulative behavior that early versions of Microsoft Bing Chat showed, despite the system having been tested extensively before launch \cite{roose2023conversation,billyperrigo,bing}.

Though there has been some progress in understanding and mitigating these issues, there is no consensus on whether or how we will be able to deeply solve them, and there is increasing concern that they will become catastrophic when exhibited in larger-scale future systems \cite{amodei2016concrete,bommasani2021CRFMreport,saunders2022critiques,ngo2022alignment}. Some experts believe that future systems trained by similar means, even if they perform well during pre-deployment testing, could fail in increasingly dramatic ways, including strategically manipulating humans to acquire power \cite{hubinger2019risks,turner2021optimal,langosco2022goal,ngo2022alignment,turner2022parametrically}. 
Broad surveys of the field suggest that these concerns are fairly broadly shared: The majority of the 738 researchers who responded to a recent survey (targeting those who published recently at the machine-learning venues NeurIPS and ICML) assigned a greater than 10\% chance of “human inability to control future advanced AI systems causing human extinction” \cite{AIimpacts2020survey}. 36\% of another sample of 480 researchers (in a survey targeting the language-specific venue ACL) agreed that “It is plausible that decisions made by AI or machine learning systems could cause a catastrophe this century that is at least as bad as an all-out nuclear war” \cite{michael2022nlp}. Hundreds of researchers recently signed a controversial open letter that calls for a moratorium on large-scale LLM training until adequate safety and governance mechanisms can be put in place \cite{fliletter}.

\section{Experts are not yet able to interpret the inner workings of LLMs}

Modern LLMs are built on artificial neural networks: They work by computing and updating numeric activation values for internal components that are very loosely modeled on human neurons \cite{bengio2017deep}. On this analogy, our tools for doing neuroscience on these systems are still weak: We have some coarse tools for testing whether models represent a few specific kinds of information (like the color results discussed in \cref{sec:LLMS_world_models}), but as of early 2023, there is no technique that would allow us to lay out in any satisfactory way what kinds of knowledge, reasoning, or goals a model is using when it produces some output. 

While there is ongoing research oriented toward this goal \citep[][i.a.]{elhage2021transformer_circuits,lovering2022unit,CausalScrubbing2022AF,burns2022latent_knowledge,li2023emergent}, the problem is deeply difficult: There are hundreds of billions of connections between these artificial neurons, some of which are invoked many times during the processing of a single piece of text, such that any attempt at a \textit{precise} explanation of an LLM’s behavior is doomed to be too complex for any human to understand. Often, ad-hoc techniques that at first seem to provide insight into the behavior of an LLM are later found to be severely misleading \cite{feng2018pathologies,jain2019attention,bolukbasi2021interpretability,wang2022towards}. In addition, promising-looking techniques that elicit reasoning in natural language do not reliably correspond to the processes that LLMs use to reason, and model-generated explanations can also be systematically misleading \cite{lipton2018mythos,ELK,uesato2022solving}.

\section{Human performance on a task isn’t an upper bound on LLM performance}

While LLMs are trained primarily to imitate human writing behavior, they can at least potentially outperform humans on many tasks. This is for two reasons: First, they are trained on far more data than any human sees,\footnote{LLMs see over $10,000\times$ more data than humans: A human adolescent sees tens of thousands of words, while LLMs can be exposed to over one trillion \cite{hart1992american,gilkerson2017mapping,hoffmann2022chinchilla}} giving them much more information to memorize and potentially synthesize. In addition, they are often given additional training using reinforcement learning before being deployed \cite{stiennon2020rlhf,ouyang2022InstructGPT,bai2022training}, which trains them to produce responses that humans find helpful \textit{without} requiring humans to demonstrate such helpful behavior. This is analogous to the techniques used to produce superhuman performance at games like Go \cite{silver2016AlphaGo}. Concretely, LLMs appear to be much better than humans at their pretraining task of predicting which word is most likely to appear after some seed piece of text \cite{Shlegeris2022AF}, and humans can teach LLMs to do some simple tasks more accurately than the humans themselves \citep{stiennon2020rlhf}.

\section{LLMs need not express the values of their creators nor the values encoded in web text\label{sec:values}}

When a plain pretrained LLM produces text, that text will generally resemble the text it was trained on. This includes a resemblance in the values expressed by the text: Models mirror their training data in the explicit statements they produce on value-laden topics and in the implicit biases behind their writing. However, these values are subject to a good degree of control by their developers, especially when the plain pretrained LLM is given further prompting and training to adapt it for deployment as a product (\cref{sec:control_and_robustness}). This means that the  values expressed in a deployed LLM's behavior do not need to reflect some average of the values expressed in its training data. This also opens up opportunities for third-party input and oversight, meaning that the values expressed in these models also need not reflect the values of the specific people and organizations who build them.

In particular, popular approaches involving reinforcement learning and \textit{red-teaming} allow model developers to guide models toward a persona and set of values more or less of their choosing \cite{dinan2019build,bai2022training,ganguli2022red}. In these techniques, the values that a model learns are never made entirely explicit. Instead, they are reflected in many small pieces of feedback that human annotators give the model during training. The \textit{constitutional AI} technique \cite{bai2022constitutional} significantly cuts down on human labor and makes these values more explicit: Using this technique, a model can be trained to follow a set of norms and values simply by writing those values down in the form of a list of constraints called a constitution. It is possible to use techniques like this to dramatically reduce \textit{explicit} examples of widely-recognized biases, like anti-Black racism, in model behavior \cite{ganguli2023capacity}.\footnote{However, explicit demonstrations of racist language or decision-making by models do not come close to exhausting the ways that the development and use of these systems interact with biases and power structures involving factors like race \citep[see, for example,][]{field2021survey}.} Indeed, in some cases, exposing models to more examples of unwanted behavior during pretraining can make it \textit{easier} to make them avoid that behavior in deployment, reversing the intuitive link between training data and model behavior (\citealt{korbak2023pretraining}; see also Appendix C in \citealt{chung2022scaling}).

These technical interventions, especially constitutional AI, are amenable to outside influence and regulation. One can easily imagine third-party standards bodies collecting input about what behaviors are acceptable in AI systems and distilling this input into constitutions that model developers are encouraged or required to adopt. 

As in \cref{sec:control_and_robustness}, though, these techniques can still fail in subtle and surprising ways, and the trends in how these techniques change as models with scale are complex. And, of course, there are many other ethical questions that arise with the development of deployment of large-scale AI systems, including issues around environmental impacts, access, misuse, privacy, safety, and the concentration of power \citep[][i.a.]{amodei2016concrete,parrots,bommasani2021CRFMreport,birhane2022values,taxonomy}.

\section{Brief interactions with LLMs are often misleading\label{sec:misleading}}

While many deployed LLMs are largely able to follow instructions, this instruction-following behavior isn't inherent to the model, but rather is grafted onto it using highly imperfect tools (\cref{sec:control_and_robustness}). In part because of this, models can be sensitive to the contents of their instructions in idiosyncratic ways. Often, a model will fail to complete a task when asked, but will then perform the task correctly once the request is reworded or reframed slightly, leading to the emerging craft of \textit{prompt engineering} \cite{brown2020gpt3,reynolds2021prompt,radford2021CLIP,dohan2022language,white2023prompt,si2023prompting}.

These contingent failures are evidence that our techniques for controlling language models to follow instructions are not reliably effective. However, simply observing that an LLM fails at a task in some setting is not reliable evidence that that LLM doesn’t have the skills or knowledge to do that task. Often, once one finds an appropriate way to prompt a model to do some task, one will find that the model \textit{consistently} performs well across different instances of the task. The chain-of-thought reasoning strategies mentioned in \cref{sec:unpredictability} are an especially clear example of this: Simply prompting a model to “think step by step” can lead it to perform well on entire categories of math and reasoning problems that it would otherwise fail on \citep{kojima2022large}. Similarly, even observing that an LLM \textit{consistently} fails at some task is far from sufficient evidence that no other LLM can do that task \citep{bowman-2022-dangers}.

On the other hand, observing that an LLM performs a task successfully in one instance is not strong evidence that the LLM is capable of performing that task in general, especially if that example was cherry-picked as part of a demonstration (like the unicorn in \cref{fig:three_unicorns}). LLMs can memorize specific examples or strategies for solving tasks from their training data \textit{without} internalizing the reasoning process that would allow them to do those tasks robustly \citep[see, e.g.][]{mccoy2019right,magar2022data}. 

\section{Discussion and Limitations}

The additional discussion that I present here builds on and contextualizes the eight claims above, but it is more speculative or subjective in places and reflects views that are not necessarily broadly shared.

\subsection{We should expect some of the prominent flaws of current LLMs to improve significantly}

Hallucination, the problem of LLMs inventing plausible false claims, is a prominent flaw in current systems and substantially limits how they can be responsibly used. Some of the recent findings discussed in \cref{sec:LLMS_world_models} suggest, though, that we may soon be able to mitigate this problem simply by finding ways to better use abilities that models already display: LLMs \textit{internally} track which statements are true with reasonably high precision, and this ability improves with scale \cite{burns2022latent_knowledge,kadavath2022language}.

Similarly, as noted in \cref{sec:values}, recent methods can dramatically reduce explicit bias and toxicity in models' output, largely by exploiting the fact that models can often recognize these bad behaviors when asked \citep{dinan2019build,bai2022constitutional,ganguli2023capacity}. While these mitigations are unlikely to be entirely robust, the prevalence and prominence of these bad behaviors will likely wane as these techniques are refined.

To be clear, though, these encouraging signs do not mean that we can reliably control these models, and the issues noted in \cref{sec:control_and_robustness} still apply. Our partial solutions are likely to leave open important failure modes. For example, straightforward attempts to manage hallucination are likely to fail silently in a way that leaves them looking more trustworthy than they are because of issues related to sandbagging: If we apply standard methods to train some future LLM to tell the truth, but that LLM can reasonably accurately predict which factual claims human data workers are likely to check, this can easily lead the LLM to tell the truth \textit{only when making claims that are likely to be checked}.

\subsection{There will be incentives to deploy LLMs as agents that flexibly pursue goals}

Increasingly capable LLMs, with increasingly accurate and usable internal models of the world, are likely to be able to take on increasingly open-ended tasks that involve making and executing novel plans to optimize for outcomes in the world \citep{chan2023harms}. As these capabilities develop, economic incentives suggest that we should see them deployed in areas like software engineering or business strategy that combine measurable outcomes, a need for flexible planning, and relatively flexible standards and regulations. LLMs augmented with additional tools can extend this into grounded domains like robotics \cite{sharma2021skill,driess2023palme}. Deployments of this type would increasingly often place LLMs in unfamiliar situations created as a result of the systems’ own actions, further reducing the degree to which their developers can predict and control their behavior. This is likely to increase the rate of simple errors that render these systems ineffective as agents in some settings. But it is also likely to increase the risk of much more dangerous errors that cause a system to remain effective while strategically pursuing the wrong goal \cite{krueger2020hidden,ortega2021shaking,chan2023harms}.

\subsection{LLM developers have limited influence over what is developed}

Because many important LLM capabilities are emergent and difficult to predict, LLM developers have relatively little influence on precisely what capabilities future LLMs will have, and efforts to make predictions about future LLM capabilities based on the economic incentives, values, or personalities of their developers are likely to fail. GPT-4, for example, appears to have many skills, like those involving programming, that its creators were likely hoping for. However, it also appears to have initially shown several unwanted skills, like teaching laypeople to prepare biological weapons, that its creators had to spend significant effort to try to remove \cite{openai2023gpt4}.

Beyond this, LLM developers inevitably also have limited \textit{awareness} of what capabilities an LLM has when they're deciding whether to deploy it: There is no known evaluation or analysis procedure that can rule out surprises like chain-of-thought reasoning in GPT-3, where users discover a way to elicit some important new behavior that the developers had not been aware of. 

\subsection{LLMs are likely to produce a rapidly growing array of risks}

More broadly, the current technical and commercial landscape provides strong incentives to build and deploy increasingly capable LLMs quickly. Nonetheless, our track record of recognizing what capabilities a new LLM can demonstrate before deploying it is spotty. Our techniques for controlling systems are weak and are likely to break down further when applied to highly capable models. Given all this, it is reasonable to expect a substantial increase and a substantial qualitative change in the range of misuse risks and model misbehaviors that emerge from the development and deployment of LLMs.

While many positive applications of LLM-based systems are likely to be genuinely valuable, the societal cost-benefit tradeoffs involved in their deployment are likely to remain difficult or impossible to evaluate in advance, at least without significant progress on hard technical problems in model evaluation, interpretability, and control. Some of these hard-to-evaluate risks, such as those involving unconventional weapons or strategic power-seeking behavior, may be impossible to adequately mitigate if they are discovered only after systems are deployed. Strategic power-seeking behavior in particular could pose serious risks during model \textit{development}, even without an intentional deployment. This suggests that future work in this area will likely warrant increasingly stringent standards for safety, security, and oversight.

\subsection{Negative results with LLMs can be difficult to interpret but point to areas of real weakness}

There are many sound scientific results showing that recent LLMs fail at language and commonsense reasoning tasks, sometimes relatively simple ones, under good-faith attempts to elicit good behavior \cite{pandia2021sorting,schuster2022sentence}. Sometimes the details of these failures cast doubts on the quality of other related evaluations \cite{webson2021prompt,ullman2023large}. For reasons mentioned in \cref{sec:misleading}, positive results on well-designed measures are much more reliable than negative results. Nonetheless, in some areas, including areas as simple as the handling of negation,\footnote{See, for example, the Modus Tollens task by Huang and Wurgaft, described in \citet{McKenzie2022Inverse}.} LLMs show what appear to be systematic weaknesses in their ability to process language or reason about the world. We have few grounds to predict whether or when these limitations will be resolved.

\subsection{The science and scholarship around LLMs is especially immature}

LLMs strain the methods and paradigms of the fields that one would expect to be best qualified to study them. Natural language processing (or language technology) is the historic home discipline for this work, but its tools are oriented toward measuring and improving the ability of computational systems to use language. While LLMs fundamentally learn and interact through language, many of the most pressing questions about their behavior and capabilities are not primarily questions about language use. The interdisciplinary fields studying AI policy and AI ethics have developed conceptual and normative frameworks for thinking about the deployment of many kinds of AI system. However, these frameworks often assume that AI systems are more precisely subject to the intentions of their human owners and developers, or to the statistics of their training data, than has been the case with recent LLMs \citep{chan2023harms}. Relatedly, many of the most cited research papers dealing with LLMs, including many papers that introduce new methods or theories, are not published in peer-reviewed venues. The recent trend toward limiting access to LLMs and treating the details of LLM training as proprietary information is also an obstacle to scientific study. 

This means that surprising novel claims about LLMs are often the product of messy, fallible science that goes beyond established disciplinary practice. However, what appears to be established conventional wisdom also often rests on shaky foundations when it is applied to LLMs. All of this is reason to be especially uncertain about the issues discussed in this paper and to make important decisions about LLMs in ways that are resilient to mistaken assumptions. 

\section*{Conclusion}

In closing, rather than recap the claims above, I would like to note three sometimes-prominent issues that this paper leaves largely untouched:

\begin{itemize}
    \item Open debates over whether we describe LLMs as \textbf{\textit{understanding}} language, and whether to describe their actions using agency-related words like \textit{know} or \textit{try}, are largely separate from the questions that I discuss here \citep[][]{bender2020climbing,michael2020dissect,potts2020possible}. We can evaluate whether systems are effective or ineffective, reliable or unreliable, interpretable or uninterpretable, and improving quickly or slowly, regardless of whether they are underlyingly human-like in the sense that these words evoke.
    \item Similarly, questions of \textbf{consciousness}, sentience, rights, and moral patienthood in LLMs \citep[see, e.g.][]{10.1093/oso/9780190905033.003.0017,shevlin_2021,chalmers2023could}, are worth distinguishing from the issues above. Though these questions may influence important decisions about how AI systems are built and used, it should be possible to evaluate most or all of the issues raised here without taking a stance on these questions.
    \item Finally, \textbf{value judgments} around LLMs are beyond the scope of this paper. The broader question of whether the rapid progress that we’re seeing with LLMs is a good thing, and what we should each do about it, depends on a deeper and more diverse range of considerations than the technical literature that I draw on here can come close to addressing. 
\end{itemize}

\section*{Acknowledgments}

This paper benefited from conversations at the AI FUTURES panel organized by Critical AI at Rutgers and from discussions with many other researchers, including Ellie Pavlick, Jackson Petty, Owain Evans, Adam Jermyn, Eric Drexler, Ben Garfinkel, Richard Ngo, Jason Wei, Helen Toner, Jeffrey Ladish, Leo Gao, Alex Lyzhov, Julian Michael, Adam Bales, Rick Korzekwa, Ben Mann, Alex Lawsen, Alex Tamkin, Anton Korinek, and David Dohan. I used LLMs in this paper only to explore some minor wording and framing decisions. All errors and omissions are, of course, my own.

This work has benefited from financial support from Eric and Wendy Schmidt (made by recommendation of the Schmidt Futures program) and from Open Philanthropy, as well as from in-kind editing support from Pablo Moreno through FAR.ai. This material is based upon work supported by the National Science Foundation under Grant Nos. 1922658 and 2046556. Any opinions, findings, and conclusions or recommendations expressed in this material are those of the author(s) and do not necessarily reflect the views of the National Science Foundation. 


\bibliography{main}

\begin{thebibliography}{120}
\providecommand{\natexlab}[1]{#1}
\providecommand{\url}[1]{\texttt{#1}}
\expandafter\ifx\csname urlstyle\endcsname\relax
  \providecommand{\doi}[1]{doi: #1}\else
  \providecommand{\doi}{doi: \begingroup \urlstyle{rm}\Url}\fi

\bibitem[Abdou et~al.(2021)Abdou, Kulmizev, Hershcovich, Frank, Pavlick, and
  S{\o}gaard]{abdou2021can}
Abdou, M., Kulmizev, A., Hershcovich, D., Frank, S., Pavlick, E., and
  S{\o}gaard, A.
\newblock Can language models encode perceptual structure without grounding? a
  case study in color.
\newblock In \emph{Proceedings of the 25th Conference on Computational Natural
  Language Learning}, pp.\  109--132, Online, November 2021. Association for
  Computational Linguistics.
\newblock \doi{10.18653/v1/2021.conll-1.9}.
\newblock URL \url{https://aclanthology.org/2021.conll-1.9}.

\bibitem[Alayrac et~al.(2022)Alayrac, Donahue, Luc, Miech, Barr, Hasson, Lenc,
  Mensch, Millican, Reynolds, et~al.]{alayrac2022flamingo}
Alayrac, J.-B., Donahue, J., Luc, P., Miech, A., Barr, I., Hasson, Y., Lenc,
  K., Mensch, A., Millican, K., Reynolds, M., et~al.
\newblock Flamingo: a visual language model for few-shot learning.
\newblock \emph{Advances in Neural Information Processing Systems},
  35:\penalty0 23716--23736, 2022.

\bibitem[Amodei et~al.(2016)Amodei, Olah, Steinhardt, Christiano, Schulman, and
  Man{\'e}]{amodei2016concrete}
Amodei, D., Olah, C., Steinhardt, J., Christiano, P., Schulman, J., and
  Man{\'e}, D.
\newblock Concrete problems in {AI} safety.
\newblock \emph{arXiv preprint 1606.06565}, 2016.

\bibitem[Andreas(2022)]{andreas2022language}
Andreas, J.
\newblock Language models as agent models.
\newblock In \emph{Findings of the Association for Computational Linguistics:
  EMNLP 2022}, pp.\  5769--5779, Abu Dhabi, United Arab Emirates, December
  2022. Association for Computational Linguistics.
\newblock URL \url{https://aclanthology.org/2022.findings-emnlp.423}.

\bibitem[Bai et~al.(2022{\natexlab{a}})Bai, Jones, Ndousse, Askell, Chen,
  DasSarma, Drain, Fort, Ganguli, Henighan, et~al.]{bai2022training}
Bai, Y., Jones, A., Ndousse, K., Askell, A., Chen, A., DasSarma, N., Drain, D.,
  Fort, S., Ganguli, D., Henighan, T., et~al.
\newblock Training a helpful and harmless assistant with reinforcement learning
  from human feedback.
\newblock \emph{arXiv preprint 2204.05862}, 2022{\natexlab{a}}.

\bibitem[Bai et~al.(2022{\natexlab{b}})Bai, Kadavath, Kundu, Askell, Kernion,
  Jones, Chen, Goldie, Mirhoseini, McKinnon, et~al.]{bai2022constitutional}
Bai, Y., Kadavath, S., Kundu, S., Askell, A., Kernion, J., Jones, A., Chen, A.,
  Goldie, A., Mirhoseini, A., McKinnon, C., et~al.
\newblock Constitutional {AI}: Harmlessness from {AI} feedback.
\newblock \emph{arXiv preprint 2212.08073}, 2022{\natexlab{b}}.

\bibitem[Bartz(2023)]{diane2023as}
Bartz, D.
\newblock As {ChatGPT's} popularity explodes, {U.S.} lawmakers take an
  interest.
\newblock \emph{Reuters}, 2023.
\newblock URL
  \url{https://www.reuters.com/technology/chatgpts-popularity-explodes-us-lawmakers-take-an-interest-2023-02-13/}.

\bibitem[Bender \& Koller(2020)Bender and Koller]{bender2020climbing}
Bender, E.~M. and Koller, A.
\newblock Climbing towards {NLU}: On meaning, form, and understanding in the
  age of data.
\newblock In \emph{Proceedings of the 58th annual meeting of the association
  for computational linguistics}, pp.\  5185--5198, 2020.

\bibitem[Bender et~al.(2021)Bender, Gebru, McMillan-Major, and
  Shmitchell]{parrots}
Bender, E.~M., Gebru, T., McMillan-Major, A., and Shmitchell, S.
\newblock On the dangers of stochastic parrots: Can language models be too big?
\newblock In \emph{Proceedings of the 2021 ACM Conference on Fairness,
  Accountability, and Transparency}, FAccT '21, pp.\  610–623, New York, NY,
  USA, 2021. Association for Computing Machinery.
\newblock ISBN 9781450383097.
\newblock \doi{10.1145/3442188.3445922}.
\newblock URL \url{https://doi.org/10.1145/3442188.3445922}.

\bibitem[Bengio et~al.(2017)Bengio, Goodfellow, and Courville]{bengio2017deep}
Bengio, Y., Goodfellow, I., and Courville, A.
\newblock \emph{Deep learning}.
\newblock MIT press Cambridge, MA, USA, 2017.
\newblock ISBN 9780262035613.

\bibitem[Bengio et~al.(2023)Bengio, Russell, Musk, Wozniak, et~al.]{fliletter}
Bengio, Y., Russell, S., Musk, E., Wozniak, S., et~al.
\newblock Pause giant {AI} experiments.
\newblock \emph{Future of Life Institute Open Letters}, 2023.
\newblock URL
  \url{https://futureoflife.org/open-letter/pause-giant-ai-experiments/}.

\bibitem[Birhane et~al.(2022)Birhane, Kalluri, Card, Agnew, Dotan, and
  Bao]{birhane2022values}
Birhane, A., Kalluri, P., Card, D., Agnew, W., Dotan, R., and Bao, M.
\newblock The values encoded in machine learning research.
\newblock In \emph{2022 ACM Conference on Fairness, Accountability, and
  Transparency}, pp.\  173--184, 2022.

\bibitem[Biswas(2023)]{biswas2023chatgpt}
Biswas, S.
\newblock {ChatGPT} and the future of medical writing.
\newblock \emph{Radiology}, pp.\  223312, 2023.

\bibitem[Bolukbasi et~al.(2021)Bolukbasi, Pearce, Yuan, Coenen, Reif,
  Vi{\'e}gas, and Wattenberg]{bolukbasi2021interpretability}
Bolukbasi, T., Pearce, A., Yuan, A., Coenen, A., Reif, E., Vi{\'e}gas, F., and
  Wattenberg, M.
\newblock An interpretability illusion for {BERT}.
\newblock \emph{arXiv preprint 2104.07143}, 2021.

\bibitem[Bommasani et~al.(2021)Bommasani, Hudson, Adeli, Altman, Arora, von
  Arx, Bernstein, Bohg, Bosselut, Brunskill, et~al.]{bommasani2021CRFMreport}
Bommasani, R., Hudson, D.~A., Adeli, E., Altman, R., Arora, S., von Arx, S.,
  Bernstein, M.~S., Bohg, J., Bosselut, A., Brunskill, E., et~al.
\newblock On the opportunities and risks of foundation models.
\newblock \emph{arXiv preprint 2108.07258}, 2021.

\bibitem[Bowman(2022)]{bowman-2022-dangers}
Bowman, S.
\newblock The dangers of underclaiming: Reasons for caution when reporting how
  {NLP} systems fail.
\newblock In \emph{Proceedings of the 60th Annual Meeting of the Association
  for Computational Linguistics (Volume 1: Long Papers)}, pp.\  7484--7499,
  Dublin, Ireland, May 2022. Association for Computational Linguistics.
\newblock \doi{10.18653/v1/2022.acl-long.516}.
\newblock URL \url{https://aclanthology.org/2022.acl-long.516}.

\bibitem[Branwen(n.d.)]{branwen}
Branwen, G.
\newblock Inner monologue ({AI}), n.d.
\newblock URL
  \url{https://gwern.net/doc/ai/nn/transformer/gpt/inner-monologue/index}.

\bibitem[Brown et~al.(2020)Brown, Mann, Ryder, Subbiah, Kaplan, Dhariwal,
  Neelakantan, Shyam, Sastry, Askell, et~al.]{brown2020gpt3}
Brown, T., Mann, B., Ryder, N., Subbiah, M., Kaplan, J.~D., Dhariwal, P.,
  Neelakantan, A., Shyam, P., Sastry, G., Askell, A., et~al.
\newblock Language models are few-shot learners.
\newblock \emph{Advances in neural information processing systems},
  33:\penalty0 1877--1901, 2020.

\bibitem[Bubeck et~al.(2023)Bubeck, Chandrasekaran, Eldan, Gehrke, Horvitz,
  Kamar, Lee, Lee, Li, Lundberg, et~al.]{bubeck2023sparks}
Bubeck, S., Chandrasekaran, V., Eldan, R., Gehrke, J., Horvitz, E., Kamar, E.,
  Lee, P., Lee, Y.~T., Li, Y., Lundberg, S., et~al.
\newblock Sparks of artificial general intelligence: Early experiments with
  {GPT-4}.
\newblock \emph{arXiv preprint 2303.12712}, 2023.

\bibitem[Burns et~al.(2023)Burns, Ye, Klein, and
  Steinhardt]{burns2022latent_knowledge}
Burns, C., Ye, H., Klein, D., and Steinhardt, J.
\newblock Discovering latent knowledge in language models without supervision.
\newblock In \emph{The Eleventh International Conference on Learning
  Representations}, 2023.
\newblock URL \url{https://openreview.net/forum?id=ETKGuby0hcs}.

\bibitem[Capoot(2023)]{ashley2023microsoft}
Capoot, A.
\newblock Microsoft announces new multibillion-dollar investment in
  {ChatGPT}-maker {OpenAI}.
\newblock \emph{CNBC}, 2023.
\newblock URL
  \url{https://www.cnbc.com/2023/01/23/microsoft-announces-multibillion-dollar-investment-in-chatgpt-maker-openai.html}.

\bibitem[Chalmers(2023)]{chalmers2023could}
Chalmers, D.~J.
\newblock Could a large language model be conscious?
\newblock \emph{arXiv preprint 2303.07103}, 2023.

\bibitem[Chan(2022)]{chan2022gpt}
Chan, A.
\newblock {GPT-3} and {InstructGPT}: technological dystopianism, utopianism,
  and ``contextual'' perspectives in {AI} ethics and industry.
\newblock \emph{{AI} and Ethics}, pp.\  1--12, 2022.

\bibitem[Chan et~al.(2023)Chan, Salganik, Markelius, Pang, Rajkumar,
  Krasheninnikov, Langosco, He, Duan, Carroll, et~al.]{chan2023harms}
Chan, A., Salganik, R., Markelius, A., Pang, C., Rajkumar, N., Krasheninnikov,
  D., Langosco, L., He, Z., Duan, Y., Carroll, M., et~al.
\newblock Harms from increasingly agentic algorithmic systems.
\newblock \emph{arXiv preprint 2302.10329}, 2023.

\bibitem[Chan et~al.(2022)Chan, Garriga-Alonso, Goldowsky-Dill, Greenblatt,
  Nitishinskaya, Radhakrishnan, Shlegeris, and Thomas]{CausalScrubbing2022AF}
Chan, L., Garriga-Alonso, A., Goldowsky-Dill, N., Greenblatt, R.,
  Nitishinskaya, J., Radhakrishnan, A., Shlegeris, B., and Thomas, N.
\newblock Causal scrubbing: a method for rigorously testing interpretability
  hypotheses.
\newblock \emph{Alignment Forum}, 2022.
\newblock URL
  \url{https://www.alignmentforum.org/posts/JvZhhzycHu2Yd57RN/causal-scrubbing-a-method-for-rigorously-testing}.

\bibitem[Choi et~al.(2023)Choi, Hickman, Monahan, and
  Schwarcz]{choi2023chatgpt}
Choi, J.~H., Hickman, K.~E., Monahan, A., and Schwarcz, D.
\newblock {ChatGPT} goes to law school.
\newblock \emph{Minnesota Legal Studies Research Paper}, 23\penalty0 (03),
  2023.
\newblock \doi{http://dx.doi.org/10.2139/ssrn.4335905}.

\bibitem[Chowdhery et~al.(2022)Chowdhery, Narang, Devlin, Bosma, Mishra,
  Roberts, Barham, Chung, Sutton, Gehrmann, et~al.]{chowdhery2022palm}
Chowdhery, A., Narang, S., Devlin, J., Bosma, M., Mishra, G., Roberts, A.,
  Barham, P., Chung, H.~W., Sutton, C., Gehrmann, S., et~al.
\newblock {PaLM}: Scaling language modeling with pathways.
\newblock \emph{arXiv preprint 2204.02311}, 2022.

\bibitem[Christiano(2022)]{ELK}
Christiano, P.
\newblock Eliciting latent knowledge.
\newblock \emph{Medium}, 2022.
\newblock URL
  \url{https://ai-alignment.com/eliciting-latent-knowledge-f977478608fc}.

\bibitem[Chung et~al.(2022)Chung, Hou, Longpre, Zoph, Tay, Fedus, Li, Wang,
  Dehghani, Brahma, et~al.]{chung2022scaling}
Chung, H.~W., Hou, L., Longpre, S., Zoph, B., Tay, Y., Fedus, W., Li, E., Wang,
  X., Dehghani, M., Brahma, S., et~al.
\newblock Scaling instruction-finetuned language models.
\newblock \emph{arXiv preprint 2210.11416}, 2022.

\bibitem[Collins et~al.(2022)Collins, Wong, Feng, Wei, and
  Tenenbaum]{collins2022structured}
Collins, K.~M., Wong, C., Feng, J., Wei, M., and Tenenbaum, J.~B.
\newblock Structured, flexible, and robust: benchmarking and improving large
  language models towards more human-like behavior in out-of-distribution
  reasoning tasks.
\newblock In \emph{2022 Cognitive Science (CogSci) conference}, 2022.

\bibitem[Devlin et~al.(2019)Devlin, Chang, Lee, and Toutanova]{devlin2018bert}
Devlin, J., Chang, M.-W., Lee, K., and Toutanova, K.
\newblock {BERT}: Pre-training of deep bidirectional transformers for language
  understanding.
\newblock In \emph{Proceedings of the 2019 Conference of the North {A}merican
  Chapter of the Association for Computational Linguistics: Human Language
  Technologies, Volume 1 (Long and Short Papers)}, pp.\  4171--4186,
  Minneapolis, Minnesota, June 2019. Association for Computational Linguistics.
\newblock \doi{10.18653/v1/N19-1423}.
\newblock URL \url{https://aclanthology.org/N19-1423}.

\bibitem[Di~Langosco et~al.(2022)Di~Langosco, Koch, Sharkey, Pfau, and
  Krueger]{langosco2022goal}
Di~Langosco, L.~L., Koch, J., Sharkey, L.~D., Pfau, J., and Krueger, D.
\newblock Goal misgeneralization in deep reinforcement learning.
\newblock In \emph{International Conference on Machine Learning}, pp.\
  12004--12019. PMLR, 2022.

\bibitem[Dinan et~al.(2019)Dinan, Humeau, Chintagunta, and
  Weston]{dinan2019build}
Dinan, E., Humeau, S., Chintagunta, B., and Weston, J.
\newblock Build it break it fix it for dialogue safety: Robustness from
  adversarial human attack.
\newblock In \emph{Proceedings of the 2019 Conference on Empirical Methods in
  Natural Language Processing and the 9th International Joint Conference on
  Natural Language Processing (EMNLP-IJCNLP)}, pp.\  4537--4546, Hong Kong,
  China, November 2019. Association for Computational Linguistics.
\newblock \doi{10.18653/v1/D19-1461}.
\newblock URL \url{https://aclanthology.org/D19-1461}.

\bibitem[Dohan et~al.(2022)Dohan, Xu, Lewkowycz, Austin, Bieber, Lopes, Wu,
  Michalewski, Saurous, Sohl-Dickstein, et~al.]{dohan2022language}
Dohan, D., Xu, W., Lewkowycz, A., Austin, J., Bieber, D., Lopes, R.~G., Wu, Y.,
  Michalewski, H., Saurous, R.~A., Sohl-Dickstein, J., et~al.
\newblock Language model cascades.
\newblock In \emph{Beyond Bayes workshop at ICML 2022}, 2022.

\bibitem[Driess et~al.(2023)Driess, Xia, Sajjadi, Lynch, Chowdhery, Ichter,
  Wahid, Tompson, Vuong, Yu, et~al.]{driess2023palme}
Driess, D., Xia, F., Sajjadi, M.~S., Lynch, C., Chowdhery, A., Ichter, B.,
  Wahid, A., Tompson, J., Vuong, Q., Yu, T., et~al.
\newblock {PaLM-E}: An embodied multimodal language model.
\newblock \emph{arXiv preprint 2303.03378}, 2023.

\bibitem[D’Amour et~al.(2020)D’Amour, Heller, Moldovan, Adlam, Alipanahi,
  Beutel, Chen, Deaton, Eisenstein, Hoffman,
  et~al.]{damour2020underspecification}
D’Amour, A., Heller, K., Moldovan, D., Adlam, B., Alipanahi, B., Beutel, A.,
  Chen, C., Deaton, J., Eisenstein, J., Hoffman, M.~D., et~al.
\newblock Underspecification presents challenges for credibility in modern
  machine learning.
\newblock \emph{Journal of Machine Learning Research}, 2020.

\bibitem[Elhage et~al.(2021)Elhage, Nanda, Olsson, Henighan, Joseph, Mann,
  Askell, Bai, Chen, Conerly, et~al.]{elhage2021transformer_circuits}
Elhage, N., Nanda, N., Olsson, C., Henighan, T., Joseph, N., Mann, B., Askell,
  A., Bai, Y., Chen, A., Conerly, T., et~al.
\newblock A mathematical framework for transformer circuits.
\newblock \emph{Transformer Circuits Thread}, 2021.
\newblock URL \url{https://transformer-circuits.pub/2021/framework/index.html}.

\bibitem[Feng et~al.(2018)Feng, Wallace, Grissom~II, Iyyer, Rodriguez, and
  Boyd-Graber]{feng2018pathologies}
Feng, S., Wallace, E., Grissom~II, A., Iyyer, M., Rodriguez, P., and
  Boyd-Graber, J.
\newblock Pathologies of neural models make interpretations difficult.
\newblock In \emph{Proceedings of the 2018 Conference on Empirical Methods in
  Natural Language Processing}, pp.\  3719--3728, Brussels, Belgium,
  October-November 2018. Association for Computational Linguistics.
\newblock \doi{10.18653/v1/D18-1407}.
\newblock URL \url{https://aclanthology.org/D18-1407}.

\bibitem[Field et~al.(2021)Field, Blodgett, Waseem, and
  Tsvetkov]{field2021survey}
Field, A., Blodgett, S.~L., Waseem, Z., and Tsvetkov, Y.
\newblock A survey of race, racism, and anti-racism in {NLP}.
\newblock In \emph{Proceedings of the 59th Annual Meeting of the Association
  for Computational Linguistics and the 11th International Joint Conference on
  Natural Language Processing (Volume 1: Long Papers)}, pp.\  1905--1925,
  Online, August 2021. Association for Computational Linguistics.
\newblock \doi{10.18653/v1/2021.acl-long.149}.
\newblock URL \url{https://aclanthology.org/2021.acl-long.149}.

\bibitem[Ganguli et~al.(2022{\natexlab{a}})Ganguli, Hernandez, Lovitt, Askell,
  Bai, Chen, Conerly, Dassarma, Drain, Elhage,
  et~al.]{ganguli2022predictability}
Ganguli, D., Hernandez, D., Lovitt, L., Askell, A., Bai, Y., Chen, A., Conerly,
  T., Dassarma, N., Drain, D., Elhage, N., et~al.
\newblock Predictability and surprise in large generative models.
\newblock In \emph{2022 ACM Conference on Fairness, Accountability, and
  Transparency}, pp.\  1747--1764, 2022{\natexlab{a}}.

\bibitem[Ganguli et~al.(2022{\natexlab{b}})Ganguli, Lovitt, Kernion, Askell,
  Bai, Kadavath, Mann, Perez, Schiefer, Ndousse, et~al.]{ganguli2022red}
Ganguli, D., Lovitt, L., Kernion, J., Askell, A., Bai, Y., Kadavath, S., Mann,
  B., Perez, E., Schiefer, N., Ndousse, K., et~al.
\newblock Red teaming language models to reduce harms: Methods, scaling
  behaviors, and lessons learned.
\newblock \emph{arXiv preprint 2209.07858}, 2022{\natexlab{b}}.

\bibitem[Ganguli et~al.(2023)Ganguli, Askell, Schiefer, Liao,
  Luko{\v{s}}i{\=u}t{\.e}, Chen, Goldie, Mirhoseini, Olsson, Hernandez,
  et~al.]{ganguli2023capacity}
Ganguli, D., Askell, A., Schiefer, N., Liao, T., Luko{\v{s}}i{\=u}t{\.e}, K.,
  Chen, A., Goldie, A., Mirhoseini, A., Olsson, C., Hernandez, D., et~al.
\newblock The capacity for moral self-correction in large language models.
\newblock \emph{arXiv preprint 2302.07459}, 2023.

\bibitem[Gilkerson et~al.(2017)Gilkerson, Richards, Warren, Montgomery,
  Greenwood, Kimbrough~Oller, Hansen, and Paul]{gilkerson2017mapping}
Gilkerson, J., Richards, J.~A., Warren, S.~F., Montgomery, J.~K., Greenwood,
  C.~R., Kimbrough~Oller, D., Hansen, J.~H., and Paul, T.~D.
\newblock Mapping the early language environment using all-day recordings and
  automated analysis.
\newblock \emph{American journal of speech-language pathology}, 26\penalty0
  (2):\penalty0 248--265, 2017.

\bibitem[Hart \& Risley(1992)Hart and Risley]{hart1992american}
Hart, B. and Risley, T.~R.
\newblock American parenting of language-learning children: Persisting
  differences in family-child interactions observed in natural home
  environments.
\newblock \emph{Developmental psychology}, 28\penalty0 (6):\penalty0 1096,
  1992.

\bibitem[He et~al.(2021)He, Liu, Gao, and Chen]{he2021deberta}
He, P., Liu, X., Gao, J., and Chen, W.
\newblock {D}e{BERT}a: Decoding-enhanced {BERT} with {D}isentangled
  {A}ttention.
\newblock In \emph{International Conference on Learning Representations}, 2021.
\newblock URL \url{https://openreview.net/forum?id=XPZIaotutsD}.

\bibitem[Hendrycks et~al.(2020)Hendrycks, Liu, Wallace, Dziedzic, Krishnan, and
  Song]{hendrycks2020pretrained}
Hendrycks, D., Liu, X., Wallace, E., Dziedzic, A., Krishnan, R., and Song, D.
\newblock Pretrained transformers improve out-of-distribution robustness.
\newblock In \emph{Proceedings of the 58th Annual Meeting of the Association
  for Computational Linguistics}, pp.\  2744--2751, Online, July 2020.
  Association for Computational Linguistics.
\newblock \doi{10.18653/v1/2020.acl-main.244}.
\newblock URL \url{https://aclanthology.org/2020.acl-main.244}.

\bibitem[Hoffmann et~al.(2022)Hoffmann, Borgeaud, Mensch, Buchatskaya, Cai,
  Rutherford, de~las Casas, Hendricks, Welbl, Clark, Hennigan, Noland,
  Millican, van~den Driessche, Damoc, Guy, Osindero, Simonyan, Elsen, Vinyals,
  Rae, and Sifre]{hoffmann2022chinchilla}
Hoffmann, J., Borgeaud, S., Mensch, A., Buchatskaya, E., Cai, T., Rutherford,
  E., de~las Casas, D., Hendricks, L.~A., Welbl, J., Clark, A., Hennigan, T.,
  Noland, E., Millican, K., van~den Driessche, G., Damoc, B., Guy, A.,
  Osindero, S., Simonyan, K., Elsen, E., Vinyals, O., Rae, J.~W., and Sifre, L.
\newblock An empirical analysis of compute-optimal large language model
  training.
\newblock In Oh, A.~H., Agarwal, A., Belgrave, D., and Cho, K. (eds.),
  \emph{Advances in Neural Information Processing Systems}, 2022.
\newblock URL \url{https://openreview.net/forum?id=iBBcRUlOAPR}.

\bibitem[Hubinger et~al.(2019)Hubinger, van Merwijk, Mikulik, Skalse, and
  Garrabrant]{hubinger2019risks}
Hubinger, E., van Merwijk, C., Mikulik, V., Skalse, J., and Garrabrant, S.
\newblock Risks from learned optimization in advanced machine learning systems.
\newblock \emph{arXiv preprint 1906.01820}, 2019.

\bibitem[J \& C(2023)J and C]{paul2023chatgpt}
J, P. and C, D.
\newblock {ChatGPT} and large language models: what's the risk?
\newblock \emph{National Cyber Security Center}, 2023.
\newblock URL
  \url{https://www.ncsc.gov.uk/blog-post/chatgpt-and-large-language-models-whats-the-risk}.

\bibitem[Jain \& Wallace(2019)Jain and Wallace]{jain2019attention}
Jain, S. and Wallace, B.~C.
\newblock {A}ttention is not {E}xplanation.
\newblock In \emph{Proceedings of the 2019 Conference of the North {A}merican
  Chapter of the Association for Computational Linguistics: Human Language
  Technologies, Volume 1 (Long and Short Papers)}, pp.\  3543--3556,
  Minneapolis, Minnesota, June 2019. Association for Computational Linguistics.
\newblock \doi{10.18653/v1/N19-1357}.
\newblock URL \url{https://aclanthology.org/N19-1357}.

\bibitem[Kadavath et~al.(2022)Kadavath, Conerly, Askell, Henighan, Drain,
  Perez, Schiefer, Dodds, DasSarma, Tran-Johnson, et~al.]{kadavath2022language}
Kadavath, S., Conerly, T., Askell, A., Henighan, T., Drain, D., Perez, E.,
  Schiefer, N., Dodds, Z.~H., DasSarma, N., Tran-Johnson, E., et~al.
\newblock Language models (mostly) know what they know.
\newblock \emph{arXiv preprint 2207.05221}, 2022.

\bibitem[Kaplan et~al.(2020)Kaplan, McCandlish, Henighan, Brown, Chess, Child,
  Gray, Radford, Wu, and Amodei]{kaplan2020scaling}
Kaplan, J., McCandlish, S., Henighan, T., Brown, T.~B., Chess, B., Child, R.,
  Gray, S., Radford, A., Wu, J., and Amodei, D.
\newblock Scaling laws for neural language models.
\newblock \emph{arXiv preprint 2001.08361}, 2020.

\bibitem[Kenton et~al.(2021)Kenton, Everitt, Weidinger, Gabriel, Mikulik, and
  Irving]{kenton2021alignment}
Kenton, Z., Everitt, T., Weidinger, L., Gabriel, I., Mikulik, V., and Irving,
  G.
\newblock Alignment of language agents.
\newblock \emph{arXiv preprint 2103.14659}, 2021.

\bibitem[Klein(2023)]{klein2023this}
Klein, E.
\newblock This changes everything.
\newblock \emph{New York Times}, 2023.
\newblock URL
  \url{https://www.nytimes.com/2023/03/12/opinion/chatbots-artificial-intelligence-future-weirdness.html}.

\bibitem[Kojima et~al.(2022)Kojima, Gu, Reid, Matsuo, and
  Iwasawa]{kojima2022large}
Kojima, T., Gu, S.~S., Reid, M., Matsuo, Y., and Iwasawa, Y.
\newblock Large language models are zero-shot reasoners.
\newblock In \emph{ICML 2022 Workshop on Knowledge Retrieval and Language
  Models}, 2022.
\newblock URL \url{https://openreview.net/forum?id=6p3AuaHAFiN}.

\bibitem[Korbak et~al.(2023)Korbak, Shi, Chen, Bhalerao, Buckley, Phang,
  Bowman, and Perez]{korbak2023pretraining}
Korbak, T., Shi, K., Chen, A., Bhalerao, R., Buckley, C.~L., Phang, J., Bowman,
  S.~R., and Perez, E.
\newblock Pretraining language models with human preferences.
\newblock \emph{arXiv preprint 2302.08582}, 2023.

\bibitem[Krueger et~al.(2020)Krueger, Maharaj, and Leike]{krueger2020hidden}
Krueger, D., Maharaj, T., and Leike, J.
\newblock Hidden incentives for auto-induced distributional shift.
\newblock \emph{arXiv preprint 2009.09153}, 2020.

\bibitem[Levesque et~al.(2012)Levesque, Davis, and
  Morgenstern]{levesque2012winograd}
Levesque, H., Davis, E., and Morgenstern, L.
\newblock The {W}inograd schema challenge.
\newblock In \emph{Thirteenth international conference on the principles of
  knowledge representation and reasoning}, 2012.

\bibitem[Li et~al.(2021)Li, Nye, and Andreas]{li2021implicit}
Li, B.~Z., Nye, M., and Andreas, J.
\newblock Implicit representations of meaning in neural language models.
\newblock In \emph{Proceedings of the 59th Annual Meeting of the Association
  for Computational Linguistics and the 11th International Joint Conference on
  Natural Language Processing (Volume 1: Long Papers)}, pp.\  1813--1827,
  Online, August 2021. Association for Computational Linguistics.
\newblock \doi{10.18653/v1/2021.acl-long.143}.
\newblock URL \url{https://aclanthology.org/2021.acl-long.143}.

\bibitem[Li et~al.(2023)Li, Hopkins, Bau, Vi{\'e}gas, Pfister, and
  Wattenberg]{li2023emergent}
Li, K., Hopkins, A.~K., Bau, D., Vi{\'e}gas, F., Pfister, H., and Wattenberg,
  M.
\newblock Emergent world representations: Exploring a sequence model trained on
  a synthetic task.
\newblock In \emph{The Eleventh International Conference on Learning
  Representations}, 2023.
\newblock URL \url{https://openreview.net/forum?id=DeG07_TcZvT}.

\bibitem[Li{\'e}tard et~al.(2021)Li{\'e}tard, Abdou, and
  S{\o}gaard]{lietard2021language}
Li{\'e}tard, B., Abdou, M., and S{\o}gaard, A.
\newblock Do language models know the way to {R}ome?
\newblock In \emph{Proceedings of the Fourth BlackboxNLP Workshop on Analyzing
  and Interpreting Neural Networks for NLP}, pp.\  510--517, Punta Cana,
  Dominican Republic, November 2021. Association for Computational Linguistics.
\newblock \doi{10.18653/v1/2021.blackboxnlp-1.40}.
\newblock URL \url{https://aclanthology.org/2021.blackboxnlp-1.40}.

\bibitem[Lieu(2023)]{lieu2023Im}
Lieu, T.
\newblock I’m a congressman who codes. {A.I.} freaks me out.
\newblock \emph{New York Times}, 2023.
\newblock URL
  \url{https://www.nytimes.com/2023/01/23/opinion/ted-lieu-ai-chatgpt-congress.html}.

\bibitem[Lipton(2018)]{lipton2018mythos}
Lipton, Z.~C.
\newblock The mythos of model interpretability: In machine learning, the
  concept of interpretability is both important and slippery.
\newblock \emph{Queue}, 16\penalty0 (3):\penalty0 31--57, 2018.

\bibitem[Lovering \& Pavlick(2022)Lovering and Pavlick]{lovering2022unit}
Lovering, C. and Pavlick, E.
\newblock Unit testing for concepts in neural networks.
\newblock \emph{Transactions of the Association for Computational Linguistics},
  10:\penalty0 1193--1208, 2022.

\bibitem[Lund \& Wang(2023)Lund and Wang]{lund2023chatting}
Lund, B.~D. and Wang, T.
\newblock Chatting about {ChatGPT}: how may {AI} and {GPT} impact academia and
  libraries?
\newblock \emph{Library Hi Tech News}, 2023.
\newblock \doi{https://doi.org/10.1108/LHTN-01-2023-0009}.

\bibitem[Magar \& Schwartz(2022)Magar and Schwartz]{magar2022data}
Magar, I. and Schwartz, R.
\newblock Data contamination: From memorization to exploitation.
\newblock In \emph{Proceedings of the 60th Annual Meeting of the Association
  for Computational Linguistics (Volume 2: Short Papers)}, pp.\  157--165,
  Dublin, Ireland, May 2022. Association for Computational Linguistics.
\newblock \doi{10.18653/v1/2022.acl-short.18}.
\newblock URL \url{https://aclanthology.org/2022.acl-short.18}.

\bibitem[McCoy et~al.(2019)McCoy, Pavlick, and Linzen]{mccoy2019right}
McCoy, T., Pavlick, E., and Linzen, T.
\newblock Right for the wrong reasons: Diagnosing syntactic heuristics in
  natural language inference.
\newblock In \emph{Proceedings of the 57th Annual Meeting of the Association
  for Computational Linguistics}, pp.\  3428--3448, Florence, Italy, July 2019.
  Association for Computational Linguistics.
\newblock \doi{10.18653/v1/P19-1334}.
\newblock URL \url{https://aclanthology.org/P19-1334}.

\bibitem[McKenzie et~al.(2022)McKenzie, Lyzhov, Pieler, Parrish, Prabhu,
  Mueller, Kim, Bowman, and Perez]{McKenzie2022Inverse}
McKenzie, I., Lyzhov, A., Pieler, M., Parrish, A., Prabhu, A., Mueller, A.,
  Kim, N., Bowman, S., and Perez, E.
\newblock Inverse scaling prize: Second round winners, 2022.
\newblock URL \url{https://irmckenzie.co.uk/round2}.

\bibitem[Mehdi(2023)]{bing}
Mehdi, Y.
\newblock Reinventing search with a new {AI}-powered {M}icrosoft {B}ing and
  {E}dge, your copilot for the web.
\newblock \emph{Official Microsoft Blog}, 2023.
\newblock URL
  \url{https://blogs.microsoft.com/blog/2023/02/07/reinventing-search-with-a-new-ai-powered-microsoft-bing-and-edge-your-copilot-for-the-web/}.

\bibitem[Menick et~al.(2022)Menick, Trebacz, Mikulik, Aslanides, Song,
  Chadwick, Glaese, Young, Campbell-Gillingham, Irving,
  et~al.]{menick2022gopher}
Menick, J., Trebacz, M., Mikulik, V., Aslanides, J., Song, F., Chadwick, M.,
  Glaese, M., Young, S., Campbell-Gillingham, L., Irving, G., et~al.
\newblock Teaching language models to support answers with verified quotes.
\newblock \emph{arXiv preprint 2203.11147}, 2022.

\bibitem[Michael(2020)]{michael2020dissect}
Michael, J.
\newblock To dissect an octopus: Making sense of the form/meaning debate.
\newblock \emph{Blog post}, 2020.
\newblock URL
  \url{https://julianmichael.org/blog/2020/07/23/to-dissect-an-octopus.html}.

\bibitem[Michael et~al.(2022)Michael, Holtzman, Parrish, Mueller, Wang, Chen,
  Madaan, Nangia, Pang, Phang, et~al.]{michael2022nlp}
Michael, J., Holtzman, A., Parrish, A., Mueller, A., Wang, A., Chen, A.,
  Madaan, D., Nangia, N., Pang, R.~Y., Phang, J., et~al.
\newblock What do {NLP} researchers believe? {R}esults of the {NLP} community
  metasurvey.
\newblock \emph{arXiv preprint 2208.12852}, 2022.

\bibitem[Nakano et~al.(2021)Nakano, Hilton, Balaji, Wu, Ouyang, Kim, Hesse,
  Jain, Kosaraju, Saunders, et~al.]{nakano2021webgpt}
Nakano, R., Hilton, J., Balaji, S., Wu, J., Ouyang, L., Kim, C., Hesse, C.,
  Jain, S., Kosaraju, V., Saunders, W., et~al.
\newblock Web{GPT}: Browser-assisted question-answering with human feedback.
\newblock \emph{arXiv preprint 2112.09332}, 2021.

\bibitem[Ngo(2022)]{ngo2022alignment}
Ngo, R.
\newblock The alignment problem from a deep learning perspective.
\newblock \emph{arXiv preprint 2209.00626}, 2022.

\bibitem[Nye et~al.(2021)Nye, Andreassen, Gur-Ari, Michalewski, Austin, Bieber,
  Dohan, Lewkowycz, Bosma, Luan, et~al.]{nye2021show}
Nye, M., Andreassen, A.~J., Gur-Ari, G., Michalewski, H., Austin, J., Bieber,
  D., Dohan, D., Lewkowycz, A., Bosma, M., Luan, D., et~al.
\newblock Show your work: Scratchpads for intermediate computation with
  language models.
\newblock \emph{arXiv preprint 2112.00114}, 2021.

\bibitem[Oliver(2023)]{lastweektonight}
Oliver, J.
\newblock Last week tonight with {J}ohn {O}liver: Feb 26, 2023.
\newblock URL
  \url{https://www.hbo.com/last-week-tonight-with-john-oliver/season-10/2-february-26-2022}.

\bibitem[OpenAI(2023{\natexlab{a}})]{openai2023ChatGPTplugins}
OpenAI.
\newblock {ChatGPT} plugins, 2023{\natexlab{a}}.
\newblock URL \url{https://openai.com/blog/chatgpt-plugins}.

\bibitem[OpenAI(2023{\natexlab{b}})]{openai2023gpt4}
OpenAI.
\newblock {GPT-4} technical report.
\newblock \emph{arXiv preprint 2303.08774}, 2023{\natexlab{b}}.
\newblock URL \url{https://doi.org/10.48550/arXiv.2303.08774}.

\bibitem[Ortega et~al.(2021)Ortega, Kunesch, Del{\'e}tang, Genewein, Grau-Moya,
  Veness, Buchli, Degrave, Piot, Perolat, et~al.]{ortega2021shaking}
Ortega, P.~A., Kunesch, M., Del{\'e}tang, G., Genewein, T., Grau-Moya, J.,
  Veness, J., Buchli, J., Degrave, J., Piot, B., Perolat, J., et~al.
\newblock Shaking the foundations: delusions in sequence models for interaction
  and control.
\newblock \emph{arXiv preprint 2110.10819}, 2021.

\bibitem[Ouyang et~al.(2022)Ouyang, Wu, Jiang, Almeida, Wainwright, Mishkin,
  Zhang, Agarwal, Slama, Ray, et~al.]{ouyang2022InstructGPT}
Ouyang, L., Wu, J., Jiang, X., Almeida, D., Wainwright, C., Mishkin, P., Zhang,
  C., Agarwal, S., Slama, K., Ray, A., et~al.
\newblock Training language models to follow instructions with human feedback.
\newblock \emph{Advances in Neural Information Processing Systems},
  35:\penalty0 27730--27744, 2022.

\bibitem[Pandia \& Ettinger(2021)Pandia and Ettinger]{pandia2021sorting}
Pandia, L. and Ettinger, A.
\newblock Sorting through the noise: Testing robustness of information
  processing in pre-trained language models.
\newblock In \emph{Conference on Empirical Methods in Natural Language
  Processing}, 2021.

\bibitem[Patel \& Pavlick(2022)Patel and Pavlick]{patel2022mapping}
Patel, R. and Pavlick, E.
\newblock Mapping language models to grounded conceptual spaces.
\newblock In \emph{International Conference on Learning Representations}, 2022.

\bibitem[Perez et~al.(2022)Perez, Ringer, Luko{\v{s}}i{\=u}t{\.e}, Nguyen,
  Chen, Heiner, Pettit, Olsson, Kundu, Kadavath, et~al.]{perez2022discovering}
Perez, E., Ringer, S., Luko{\v{s}}i{\=u}t{\.e}, K., Nguyen, K., Chen, E.,
  Heiner, S., Pettit, C., Olsson, C., Kundu, S., Kadavath, S., et~al.
\newblock Discovering language model behaviors with model-written evaluations.
\newblock \emph{arXiv preprint 2212.09251}, 2022.

\bibitem[Perrigo(2023)]{billyperrigo}
Perrigo, B.
\newblock The new {AI}-powered {B}ing is threatening users. that’s no
  laughing matter.
\newblock \emph{Time}, 2023.
\newblock URL
  \url{https://time.com/6256529/bing-openai-chatgpt-danger-alignment/}.

\bibitem[Potts(2020)]{potts2020possible}
Potts, C.
\newblock Is it possible for language models to achieve language understanding.
\newblock \emph{Medium}, 2020.
\newblock URL
  \url{https://chrisgpotts.medium.com/is-it-possible-for-language-models-to-achieve-language-understanding-81df45082ee2}.

\bibitem[Radford et~al.(2018)Radford, Narasimhan, Salimans, Sutskever,
  et~al.]{radford2018gpt1}
Radford, A., Narasimhan, K., Salimans, T., Sutskever, I., et~al.
\newblock Improving language understanding by generative pre-training.
\newblock \emph{OpenAI blog}, 2018.
\newblock URL \url{https://openai.com/research/language-unsupervised}.

\bibitem[Radford et~al.(2019)Radford, Wu, Child, Luan, Amodei, Sutskever,
  et~al.]{radford2019gpt2}
Radford, A., Wu, J., Child, R., Luan, D., Amodei, D., Sutskever, I., et~al.
\newblock Language models are unsupervised multitask learners.
\newblock \emph{OpenAI blog}, 2019.
\newblock URL \url{https://openai.com/research/better-language-models}.

\bibitem[Radford et~al.(2021)Radford, Kim, Hallacy, Ramesh, Goh, Agarwal,
  Sastry, Askell, Mishkin, Clark, et~al.]{radford2021CLIP}
Radford, A., Kim, J.~W., Hallacy, C., Ramesh, A., Goh, G., Agarwal, S., Sastry,
  G., Askell, A., Mishkin, P., Clark, J., et~al.
\newblock Learning transferable visual models from natural language
  supervision.
\newblock In \emph{International conference on machine learning}, pp.\
  8748--8763. PMLR, 2021.

\bibitem[Raffel et~al.(2020)Raffel, Shazeer, Roberts, Lee, Narang, Matena,
  Zhou, Li, and Liu]{raffel2020T5}
Raffel, C., Shazeer, N., Roberts, A., Lee, K., Narang, S., Matena, M., Zhou,
  Y., Li, W., and Liu, P.~J.
\newblock Exploring the limits of transfer learning with a unified text-to-text
  transformer.
\newblock \emph{The Journal of Machine Learning Research}, 21\penalty0
  (1):\penalty0 5485--5551, 2020.

\bibitem[Reynolds \& McDonell(2021)Reynolds and McDonell]{reynolds2021prompt}
Reynolds, L. and McDonell, K.
\newblock Prompt programming for large language models: Beyond the few-shot
  paradigm.
\newblock In \emph{Extended Abstracts of the 2021 CHI Conference on Human
  Factors in Computing Systems}, pp.\  1--7, 2021.

\bibitem[Roose(2023)]{roose2023conversation}
Roose, K.
\newblock A conversation with {B}ing’s chatbot left me deeply unsettled.
\newblock \emph{New York Times}, 2023.
\newblock URL
  \url{https://www.nytimes.com/2023/02/16/technology/bing-chatbot-microsoft-chatgpt.html}.

\bibitem[Saunders et~al.(2022)Saunders, Yeh, Wu, Bills, Ouyang, Ward, and
  Leike]{saunders2022critiques}
Saunders, W., Yeh, C., Wu, J., Bills, S., Ouyang, L., Ward, J., and Leike, J.
\newblock Self-critiquing models for assisting human evaluators.
\newblock \emph{arXiv preprint 2206.05802}, 2022.

\bibitem[Schick et~al.(2023)Schick, Dwivedi-Yu, Dess{\`\i}, Raileanu, Lomeli,
  Zettlemoyer, Cancedda, and Scialom]{schick2023toolformer}
Schick, T., Dwivedi-Yu, J., Dess{\`\i}, R., Raileanu, R., Lomeli, M.,
  Zettlemoyer, L., Cancedda, N., and Scialom, T.
\newblock Toolformer: Language models can teach themselves to use tools.
\newblock \emph{arXiv preprint 2302.04761}, 2023.

\bibitem[Schuster \& Linzen(2022)Schuster and Linzen]{schuster2022sentence}
Schuster, S. and Linzen, T.
\newblock When a sentence does not introduce a discourse entity,
  transformer-based models still sometimes refer to it.
\newblock In \emph{Proceedings of the 2022 Conference of the North American
  Chapter of the Association for Computational Linguistics: Human Language
  Technologies}, pp.\  969--982, Seattle, United States, July 2022. Association
  for Computational Linguistics.
\newblock \doi{10.18653/v1/2022.naacl-main.71}.
\newblock URL \url{https://aclanthology.org/2022.naacl-main.71}.

\bibitem[Schwitzgebel \& Garza(2020)Schwitzgebel and
  Garza]{10.1093/oso/9780190905033.003.0017}
Schwitzgebel, E. and Garza, M.
\newblock {Designing {AI} with Rights, Consciousness, Self-Respect, and
  Freedom}.
\newblock In \emph{{Ethics of Artificial Intelligence}}. Oxford University
  Press, 09 2020.
\newblock ISBN 9780190905033.
\newblock \doi{10.1093/oso/9780190905033.003.0017}.
\newblock URL \url{https://doi.org/10.1093/oso/9780190905033.003.0017}.

\bibitem[Sevilla et~al.(2022)Sevilla, Heim, Ho, Besiroglu, Hobbhahn, and
  Villalobos]{sevilla2022epoch}
Sevilla, J., Heim, L., Ho, A., Besiroglu, T., Hobbhahn, M., and Villalobos, P.
\newblock Compute trends across three eras of machine learning.
\newblock In \emph{2022 International Joint Conference on Neural Networks
  (IJCNN)}, pp.\  1--8. IEEE, 2022.

\bibitem[Sharma et~al.(2022)Sharma, Torralba, and Andreas]{sharma2021skill}
Sharma, P., Torralba, A., and Andreas, J.
\newblock Skill induction and planning with latent language.
\newblock In \emph{Proceedings of the 60th Annual Meeting of the Association
  for Computational Linguistics (Volume 1: Long Papers)}, pp.\  1713--1726,
  Dublin, Ireland, May 2022. Association for Computational Linguistics.
\newblock \doi{10.18653/v1/2022.acl-long.120}.
\newblock URL \url{https://aclanthology.org/2022.acl-long.120}.

\bibitem[Shevlin(2021)]{shevlin_2021}
Shevlin, H.
\newblock How could we know when a robot was a moral patient?
\newblock \emph{Cambridge Quarterly of Healthcare Ethics}, 30\penalty0
  (3):\penalty0 459–471, 2021.
\newblock \doi{10.1017/S0963180120001012}.

\bibitem[Shlegeris et~al.(2022)Shlegeris, Roger, and Chan]{Shlegeris2022AF}
Shlegeris, B., Roger, F., and Chan, L.
\newblock Language models seem to be much better than humans at next-token
  prediction.
\newblock \emph{Alignment Forum}, 2022.
\newblock URL
  \url{https://www.alignmentforum.org/posts/htrZrxduciZ5QaCjw/language-models-seem-to-be-much-better-than-humans-at-next}.

\bibitem[Si et~al.(2023)Si, Gan, Yang, Wang, Wang, Boyd-Graber, and
  Wang]{si2023prompting}
Si, C., Gan, Z., Yang, Z., Wang, S., Wang, J., Boyd-Graber, J.~L., and Wang, L.
\newblock Prompting {GPT}-3 to be reliable.
\newblock In \emph{The Eleventh International Conference on Learning
  Representations}, 2023.
\newblock URL \url{https://openreview.net/forum?id=98p5x51L5af}.

\bibitem[Silver et~al.(2016)Silver, Huang, Maddison, Guez, Sifre, Van
  Den~Driessche, Schrittwieser, Antonoglou, Panneershelvam, Lanctot,
  et~al.]{silver2016AlphaGo}
Silver, D., Huang, A., Maddison, C.~J., Guez, A., Sifre, L., Van Den~Driessche,
  G., Schrittwieser, J., Antonoglou, I., Panneershelvam, V., Lanctot, M.,
  et~al.
\newblock Mastering the game of {G}o with deep neural networks and tree search.
\newblock \emph{Nature}, 529\penalty0 (7587):\penalty0 484--489, 2016.

\bibitem[S{\o}gaard(2023)]{sogaard2023grounding}
S{\o}gaard, A.
\newblock Grounding the vector space of an octopus: Word meaning from raw text.
\newblock \emph{Minds and Machines}, pp.\  1--22, 2023.

\bibitem[Srivastava et~al.(2022)Srivastava, Rastogi, Rao, Shoeb, Abid, Fisch,
  Brown, Santoro, Gupta, Garriga-Alonso, et~al.]{srivastava2022beyond}
Srivastava, A., Rastogi, A., Rao, A., Shoeb, A. A.~M., Abid, A., Fisch, A.,
  Brown, A.~R., Santoro, A., Gupta, A., Garriga-Alonso, A., et~al.
\newblock Beyond the imitation game: Quantifying and extrapolating the
  capabilities of language models.
\newblock \emph{arXiv preprint 2206.04615}, 2022.

\bibitem[Stein-Perlman et~al.(2020)Stein-Perlman, Weinstein-Raun, and
  Grace]{AIimpacts2020survey}
Stein-Perlman, Z., Weinstein-Raun, B., and Grace, K.
\newblock 2022 expert survey on progress in {AI}.
\newblock \emph{{AI} Impacts blog}, 2020.
\newblock URL
  \url{https://aiimpacts.org/2022-expert-survey-on-progress-in-ai/}.

\bibitem[Steinhardt(2021)]{Steinhardt2021risks}
Steinhardt, J.
\newblock On the risks of emergent behavior in foundation models.
\newblock \emph{Stanford CRFM blog post}, 2021.
\newblock URL
  \url{https://crfm.stanford.edu/commentary/2021/10/18/steinhardt.html}.

\bibitem[Steinhardt(2022)]{Steinhardt2021one}
Steinhardt, J.
\newblock {AI} forecasting: One year in.
\newblock \emph{Bounded Regret}, 2022.
\newblock URL
  \url{https://bounded-regret.ghost.io/ai-forecasting-one-year-in/}.

\bibitem[Stiennon et~al.(2020)Stiennon, Ouyang, Wu, Ziegler, Lowe, Voss,
  Radford, Amodei, and Christiano]{stiennon2020rlhf}
Stiennon, N., Ouyang, L., Wu, J., Ziegler, D., Lowe, R., Voss, C., Radford, A.,
  Amodei, D., and Christiano, P.~F.
\newblock Learning to summarize with human feedback.
\newblock \emph{Advances in Neural Information Processing Systems},
  33:\penalty0 3008--3021, 2020.

\bibitem[Touvron et~al.(2023)Touvron, Lavril, Izacard, Martinet, Lachaux,
  Lacroix, Rozi{\`e}re, Goyal, Hambro, Azhar, et~al.]{touvron2023llama}
Touvron, H., Lavril, T., Izacard, G., Martinet, X., Lachaux, M.-A., Lacroix,
  T., Rozi{\`e}re, B., Goyal, N., Hambro, E., Azhar, F., et~al.
\newblock {LLaMA}: Open and efficient foundation language models.
\newblock \emph{arXiv preprint 2302.13971}, 2023.

\bibitem[Turner \& Tadepalli(2022)Turner and
  Tadepalli]{turner2022parametrically}
Turner, A. and Tadepalli, P.
\newblock Parametrically retargetable decision-makers tend to seek power.
\newblock \emph{Advances in Neural Information Processing Systems},
  35:\penalty0 31391--31401, 2022.

\bibitem[Turner et~al.(2021)Turner, Smith, Shah, Critch, and
  Tadepalli]{turner2021optimal}
Turner, A.~M., Smith, L.~R., Shah, R., Critch, A., and Tadepalli, P.
\newblock Optimal policies tend to seek power.
\newblock In Beygelzimer, A., Dauphin, Y., Liang, P., and Vaughan, J.~W.
  (eds.), \emph{Advances in Neural Information Processing Systems}, 2021.
\newblock URL \url{https://openreview.net/forum?id=l7-DBWawSZH}.

\bibitem[Uesato et~al.(2022)Uesato, Kushman, Kumar, Song, Siegel, Wang,
  Creswell, Irving, and Higgins]{uesato2022solving}
Uesato, J., Kushman, N., Kumar, R., Song, F., Siegel, N., Wang, L., Creswell,
  A., Irving, G., and Higgins, I.
\newblock Solving math word problems with process-and outcome-based feedback.
\newblock \emph{arXiv preprint 2211.14275}, 2022.

\bibitem[Ullman(2023)]{ullman2023large}
Ullman, T.
\newblock Large language models fail on trivial alterations to theory-of-mind
  tasks.
\newblock \emph{arXiv preprint 2302.08399}, 2023.

\bibitem[Wang et~al.(2022)Wang, Min, Deng, Shen, Wu, Zettlemoyer, and
  Sun]{wang2022towards}
Wang, B., Min, S., Deng, X., Shen, J., Wu, Y., Zettlemoyer, L., and Sun, H.
\newblock Towards understanding chain-of-thought prompting: An empirical study
  of what matters.
\newblock \emph{arXiv preprint 2212.10001}, 2022.

\bibitem[Webson \& Pavlick(2022)Webson and Pavlick]{webson2021prompt}
Webson, A. and Pavlick, E.
\newblock Do prompt-based models really understand the meaning of their
  prompts?
\newblock In \emph{Proceedings of the 2022 Conference of the North American
  Chapter of the Association for Computational Linguistics: Human Language
  Technologies}, pp.\  2300--2344, Seattle, United States, July 2022.
  Association for Computational Linguistics.
\newblock \doi{10.18653/v1/2022.naacl-main.167}.
\newblock URL \url{https://aclanthology.org/2022.naacl-main.167}.

\bibitem[Wei et~al.(2022{\natexlab{a}})Wei, Tay, Bommasani, Raffel, Zoph,
  Borgeaud, Yogatama, Bosma, Zhou, Metzler, Chi, Hashimoto, Vinyals, Liang,
  Dean, and Fedus]{wei2022emergent}
Wei, J., Tay, Y., Bommasani, R., Raffel, C., Zoph, B., Borgeaud, S., Yogatama,
  D., Bosma, M., Zhou, D., Metzler, D., Chi, E.~H., Hashimoto, T., Vinyals, O.,
  Liang, P., Dean, J., and Fedus, W.
\newblock Emergent abilities of large language models.
\newblock \emph{Transactions on Machine Learning Research}, 2022{\natexlab{a}}.
\newblock ISSN 2835-8856.
\newblock URL \url{https://openreview.net/forum?id=yzkSU5zdwD}.
\newblock Survey Certification.

\bibitem[Wei et~al.(2022{\natexlab{b}})Wei, Wang, Schuurmans, Bosma, brian
  ichter, Xia, Chi, Le, and Zhou]{wei2022chain}
Wei, J., Wang, X., Schuurmans, D., Bosma, M., brian ichter, Xia, F., Chi,
  E.~H., Le, Q.~V., and Zhou, D.
\newblock Chain of thought prompting elicits reasoning in large language
  models.
\newblock In Oh, A.~H., Agarwal, A., Belgrave, D., and Cho, K. (eds.),
  \emph{Advances in Neural Information Processing Systems}, 2022{\natexlab{b}}.
\newblock URL \url{https://openreview.net/forum?id=_VjQlMeSB_J}.

\bibitem[Weidinger et~al.(2022)Weidinger, Uesato, Rauh, Griffin, Huang, Mellor,
  Glaese, Cheng, Balle, Kasirzadeh, Biles, Brown, Kenton, Hawkins, Stepleton,
  Birhane, Hendricks, Rimell, Isaac, Haas, Legassick, Irving, and
  Gabriel]{taxonomy}
Weidinger, L., Uesato, J., Rauh, M., Griffin, C., Huang, P.-S., Mellor, J.,
  Glaese, A., Cheng, M., Balle, B., Kasirzadeh, A., Biles, C., Brown, S.,
  Kenton, Z., Hawkins, W., Stepleton, T., Birhane, A., Hendricks, L.~A.,
  Rimell, L., Isaac, W., Haas, J., Legassick, S., Irving, G., and Gabriel, I.
\newblock Taxonomy of risks posed by language models.
\newblock In \emph{2022 ACM Conference on Fairness, Accountability, and
  Transparency}, FAccT '22, pp.\  214–229, New York, NY, USA, 2022.
  Association for Computing Machinery.
\newblock ISBN 9781450393522.
\newblock \doi{10.1145/3531146.3533088}.
\newblock URL \url{https://doi.org/10.1145/3531146.3533088}.

\bibitem[White et~al.(2023)White, Fu, Hays, Sandborn, Olea, Gilbert, Elnashar,
  Spencer-Smith, and Schmidt]{white2023prompt}
White, J., Fu, Q., Hays, S., Sandborn, M., Olea, C., Gilbert, H., Elnashar, A.,
  Spencer-Smith, J., and Schmidt, D.~C.
\newblock A prompt pattern catalog to enhance prompt engineering with
  {ChatGPT}.
\newblock \emph{arXiv preprint 2302.11382}, 2023.

\bibitem[Zhou et~al.(2023)Zhou, Sch{\"a}rli, Hou, Wei, Scales, Wang,
  Schuurmans, Cui, Bousquet, Le, and Chi]{zhou2023least}
Zhou, D., Sch{\"a}rli, N., Hou, L., Wei, J., Scales, N., Wang, X., Schuurmans,
  D., Cui, C., Bousquet, O., Le, Q.~V., and Chi, E.~H.
\newblock Least-to-most prompting enables complex reasoning in large language
  models.
\newblock In \emph{The Eleventh International Conference on Learning
  Representations}, 2023.
\newblock URL \url{https://openreview.net/forum?id=WZH7099tgfM}.

\bibitem[Ziegler et~al.(2019)Ziegler, Stiennon, Wu, Brown, Radford, Amodei,
  Christiano, and Irving]{ziegler2019fine}
Ziegler, D.~M., Stiennon, N., Wu, J., Brown, T.~B., Radford, A., Amodei, D.,
  Christiano, P., and Irving, G.
\newblock Fine-tuning language models from human preferences.
\newblock \emph{arXiv preprint 1909.08593}, 2019.

\end{thebibliography}
\bibliographystyle{icml2023}


\end{document}